\documentclass[runningheads]{llncs}

\usepackage{eccv}

\usepackage{subcaption}
\usepackage{booktabs}
\usepackage{multicol}
\usepackage{multirow}
\usepackage{wrapfig}
\usepackage{siunitx}
\usepackage{graphicx}
\usepackage{xcolor}   
\usepackage{makecell} 
\usepackage{algorithm}
\usepackage{algpseudocode}
\usepackage{amsmath}
\usepackage{amssymb}
\usepackage{arydshln}
\usepackage{caption}
\usepackage{eso-pic}

\usepackage[table,xcdraw]{xcolor}
\usepackage{eccvabbrv}

\usepackage{graphicx}
\usepackage{booktabs}

\usepackage[accsupp]{axessibility}  

\usepackage{hyperref}

\usepackage{orcidlink}

\begin{document}

\title{Preserving Knowledge across Space and Time \\for Continual Video Deepfake Detection} 
\titlerunning{Preserving Knowledge across Space and Time}


\author{
Taehoon Kim\inst{1}$^{\star}$\orcidlink{0009-0000-1474-9870} \and
Jongwook Choi\inst{1}$^{\star}$\orcidlink{0009-0009-1844-3292} \and
Heejae Jo\inst{2}\orcidlink{0009-0001-3194-6582} \and \\
Byungmin Park\inst{1}\orcidlink{0009-0009-1577-9268} \and 
Jongwon Choi\inst{1,2,3}$^{\dagger}$\orcidlink{0000-0001-9753-8760}
}

\authorrunning{T.~Kim et al.}

\institute{
GS. of AI, Chung-Ang University, Republic of Korea \and
Dept. of Advanced Imaging, GSAIM, Chung-Ang University, Republic of Korea \and
Dept. of Metaverse Convergence, Chung-Ang University, Republic of Korea \\
\email{\{kimth,cjw,jzoe,byungmin\_park\}@vilab.cau.ac.kr, choijw@cau.ac.kr}
}


\maketitle
\AddToShipoutPictureFG*{%
  \AtPageUpperLeft{%
    \put(\LenToUnit{2.0cm},\LenToUnit{-1.2cm}){%
      \makebox(0,0)[lt]{%
        \footnotesize\itshape
        \begin{tabular}{@{}l@{}}
          Preprint version; final version will be available at the European Conference on Computer Vision (ECCV) (2026).\\
          Code will be available at
          \href{https://github.com/rama0126/MSFD}{github.com/rama0126/MSFD}.
        \end{tabular}%
      }%
    }%
  }%
}
\renewcommand{\thefootnote}{}
\footnotetext{$^{\star}$ Equal contribution. \quad $^{\dagger}$ Corresponding author.}
\renewcommand{\thefootnote}{\arabic{footnote}}

\begin{abstract}
The continuous emergence of high-quality video deepfakes requires detectors that continually adapt to new forgery patterns, yet existing approaches, which are designed for deepfake images, fail to capture video-specific cues.
Unlike deepfake images that contain only spatial artifacts, deepfake videos leave distinct evidence along both spatial and temporal axes, necessitating the separate preservation of each modality during sequential model updates. 
To overcome this limitation, we introduce a continual deepfake video detection framework, Modality-Specific Frequency Distillation (MSFD), that explicitly decomposes video features into spatial, temporal, and spatiotemporal modalities in the frequency domain. 
This decomposition enables independent preservation of each modality, as different deepfake video types exhibit varying reliance on spatial and temporal cues across tasks.
Furthermore, MSFD adopts a cross-modality decorrelation loss that encourages spatiotemporal representations to remain orthogonal to single-modality cues.
Extensive experiments show that our framework achieves stronger adaptation and preserves performance more effectively than state-of-the-art methods across diverse continual deepfake video scenarios.
\end{abstract}

\section{Introduction}
\label{sec:intro}


To address the continuously evolving video deepfakes, continual learning approaches have been proposed to enable detection models to adapt to newly emerging forgery techniques while preserving their ability to detect previously learned ones~\cite{cored,surlid, pan2023dfil}.
The rapid advancement of deepfake generation techniques has led to the continuous emergence of new video forgery methods, with the quality and realism of generated videos improving significantly~\cite{DF40}. 
Recent video generation approaches~\cite{svd, wang2025lavie, zhang2025magicmirror} produce high-quality video deepfakes that pose substantial challenges to deepfake detection systems trained on limited forgery types.
While recent video deepfake detectors improve robustness by modeling generalizable spatiotemporal representations~\cite{ftcn, altfreezing, cvpr25_dfd_fcg}, they are typically trained once under a fixed distribution and evaluated in a static setting. 
In practice, however, forgery techniques evolve after deployment and are encountered sequentially, requiring detectors to be updated over time.

\begin{figure}[t!]
\centering
\includegraphics[width=\linewidth]{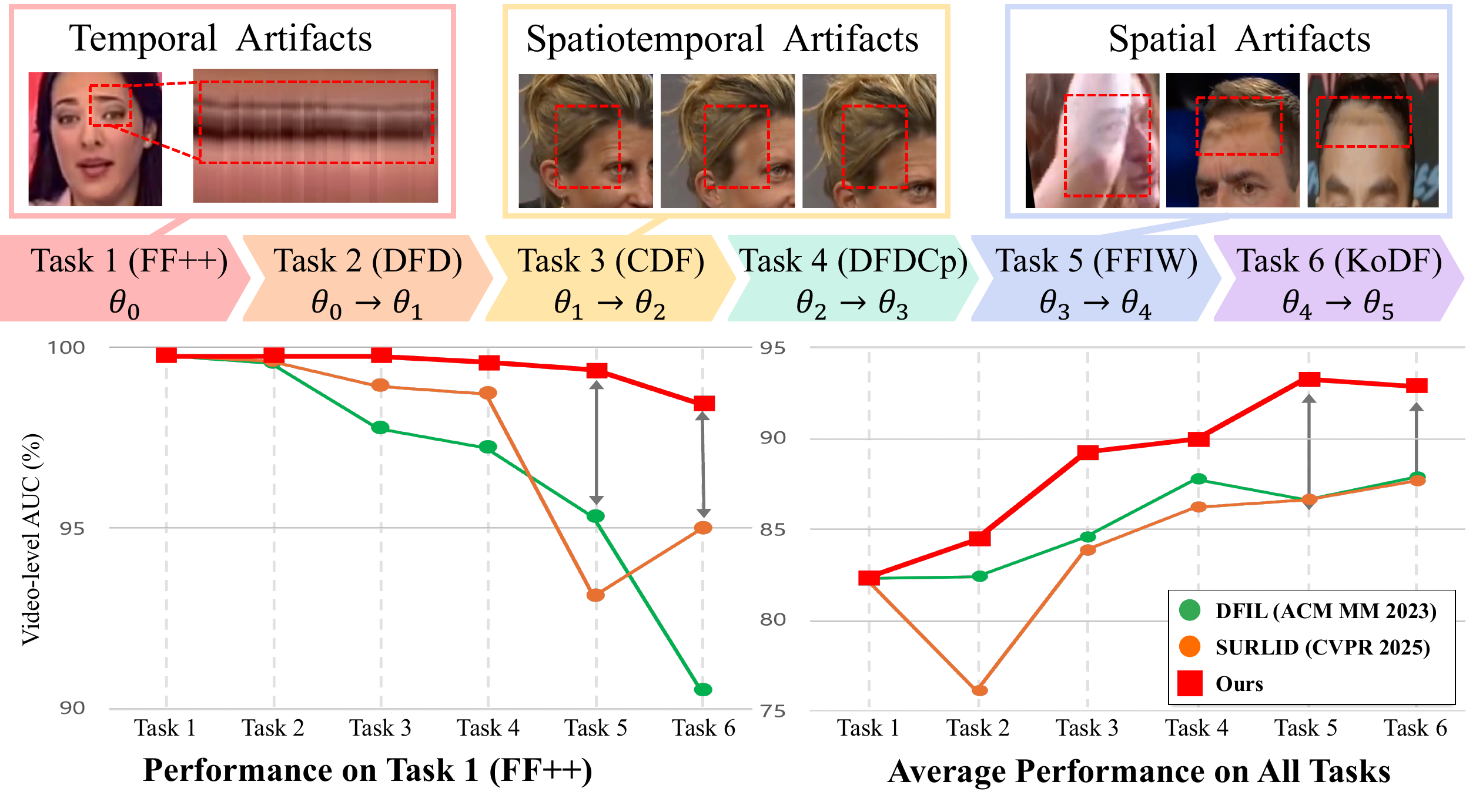} 
\caption{\textbf{Performance in the continual video deepfake detection.}
The top row illustrates representative temporal, spatiotemporal, and spatial artifacts across sequential tasks. 
The bottom row shows the performance degradation on the initial task (FF++) and average performance across all tasks as the model is continually trained on six sequential deepfake video datasets (FF++, DFD, CDF, DFDCp, FFIW, and KoDF).
Our method (red) maintains the highest performance with the least degradation compared to existing continual deepfake detection approaches.
}
\label{fig:teaser}
\end{figure}

However, most existing continual deepfake detection frameworks~\cite{cored,surlid,pan2023dfil,tian2024dynamic,ma2024hidd} were primarily designed under image-level assumptions, which limit their ability to capture video-level spatiotemporal artifacts.
Deepfake videos have complementary but distinct evidence in spatial artifacts (\eg, blending boundaries) and temporal inconsistencies (\eg, flickering, unnatural motion)~\cite{ftcn, altfreezing, nguyen2025vulnerability}.
Such image-centric designs are therefore inadequate for videos, where deepfakes exhibit characteristics that differ from those observed in manipulated images~\cite{ftcn, altfreezing}. Consequently, these image-based approaches struggle when applied to video-based detection, often leading to unbalanced adaptation and severe forgetting, as shown in Fig.~\ref{fig:teaser}.

\begin{figure}[t!]
\centering
\includegraphics[width=\linewidth]{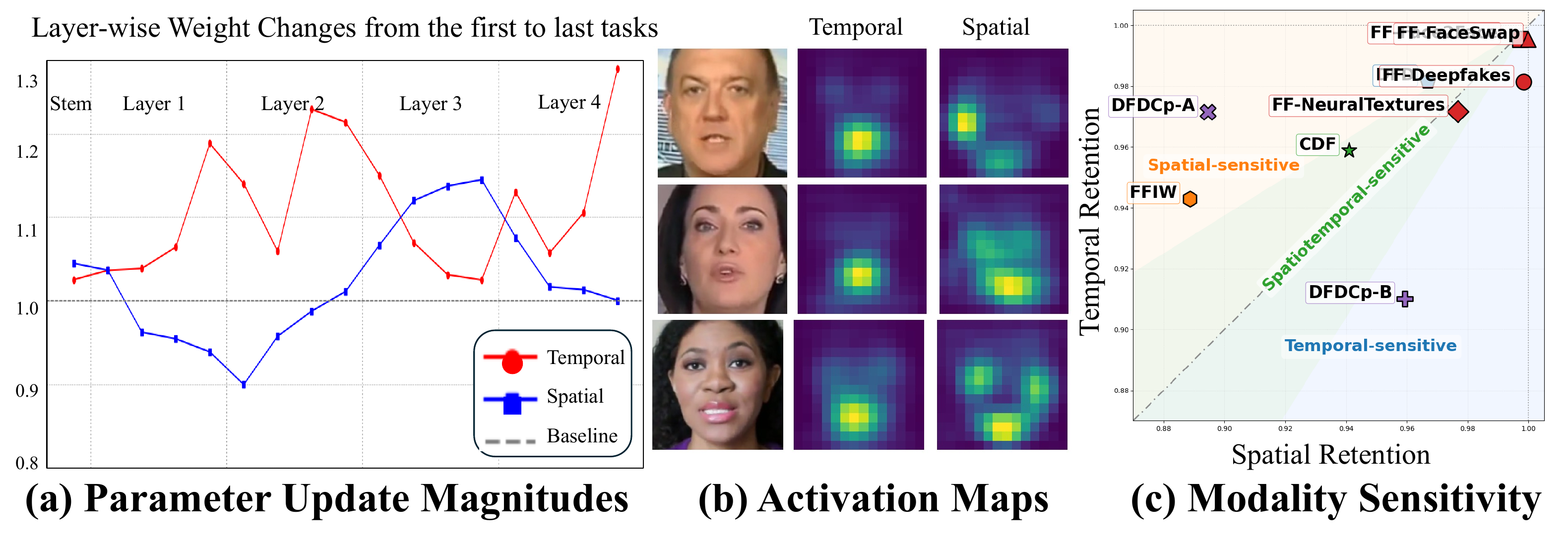}
\caption{\textbf{Analysis of Spatial and Temporal Modalities.}
We visualize (a) parameter update magnitudes across layers relative to the iCaRL~\cite{rebuffi2017icarl} baseline and (b) activation patterns when models are continually trained with either spatial-only or temporal-only cues, alongside (c) a dataset-wise sensitivity map showing the AUC retention ratio of dataset-specific finetuned models under spatial and temporal perturbations. Detailed settings are provided in the supplementary material.
}
\label{fig:weightdiffer}
\end{figure}

Our preliminary experiments further support this observation (Fig.~\ref{fig:weightdiffer}).
When we trained models using only spatial or only temporal cues in a continual-training setting, their layer-wise weight updates and activation responses diverged substantially (Fig.~\ref{fig:weightdiffer} (a) and (b)).
Moreover, performance retention under spatial and temporal perturbations varies markedly across generation paradigms—some rely primarily on spatial cues, others on temporal cues, and some on both—even within the same dataset (\eg, DFDCp-A vs.\ DFDCp-B) (Fig.~\ref{fig:weightdiffer} (c)).
These results explain why existing methods that treat spatial and temporal cues as a single entangled representation fail to retain both during continual updates, and motivate the core design of our framework: preserving spatial, temporal, and spatiotemporal representations separately.

Motivated by these observations, we develop Modality-Specific Frequency Distillation (MSFD), which decomposes features into spatial, temporal, and spatiotemporal modalities and preserves each by aligning the current model's frequency spectra with those of the previous model at each continual learning step.
This design enables independent preservation of each modality, as different deepfake video types carry distinct spatial and temporal characteristics that need to be retained separately (Fig.~\ref{fig:weightdiffer}~(c)).
We perform this decomposition in the spectral domain because it offers a simple yet highly effective approach specifically for deepfakes. 
Unlike complex disentanglement in the raw feature space, frequency transforms explicitly isolate distinct forgery signatures: spatial forgeries manifest high-frequency boundary artifacts captured by 2D spatial spectra~\cite{highfreq, bihpf}, while point-wise 1D temporal frequency spectra capture fine-grained temporal irregularities characteristic of temporal manipulations~\cite{pwtf}.
We further employ a 3D spatiotemporal spectrum to capture joint space-time dynamics inaccessible to single-modality transforms, and introduce a cross-modality decorrelation loss that encourages the spatiotemporal representation to encode frequency patterns specific to the joint use of spatial and temporal cues while remaining orthogonal to those captured by single-modality branches.

Through extensive experiments across realistic continual deepfake video detection scenarios, including sequential domain shifts, emerging forgery techniques, and restricted data, our method shows strong adaptability while preserving prior knowledge. We further confirm its robustness through evaluations on diverse perturbation settings, and our ablation studies validate the effectiveness of each component in the modality-specific design.
Our contributions are summarized as follows:
\begin{itemize}
    \item We introduce the first framework that addresses the preservation of spatial and temporal forgery cues for continual deepfake video detection.
    \item We design the Modality-Specific Frequency Distillation (MSFD) framework, which decomposes video features into three complementary frequency modalities to preserve modality-specific deepfake cues.
    \item We propose a Cross-Modality Decorrelation Loss (CDL) that encourages spatiotemporal frequency representations orthogonal to single-modality cues.
    \item We conduct comprehensive experiments across multiple continual deepfake video settings, demonstrating our proposed framework's ability to improve generalization, robustness, and forgetting mitigation.
\end{itemize}

\section{Related Work}
\label{sec:related}
\noindent{\textbf{Video-based Deepfake Detection.}}
Early deepfake detectors~\cite{Xception, highfreq, bihpf, yan2023ucf, frepgan, wang2020cnn, sbi} focused on spatial artifacts in static images, failing to account for temporal inconsistencies that often expose video manipulations.
Later studies~\cite{ftcn, lips, realforensics, altfreezing, StyleFlow, twobranchrnn, pwtf, xu2023tall, STa, nguyen2025vulnerability} showed that modeling spatiotemporal representations improves robustness to unseen forgery types.
For example, FTCN~\cite{ftcn} leverages temporal artifacts as transferable cues, while AltFreezing~\cite{altfreezing} disentangles spatial and temporal branches to stabilize detection, and recent detectors further highlight the importance of balanced spatiotemporal fusion~\cite{cvpr25_dfd_fcg, nguyen2025vulnerability, STa}.
Despite these advances, most detectors still assume a fixed closed-world distribution. 
However, new deepfake techniques are continuously evolving after deployment, necessitating detectors that can be incrementally updated over time.

\noindent{\textbf{Continual Deepfake Detection.}}
Continual deepfake detection aims to adapt detectors to newly emerging forgeries while retaining performance on previously learned ones~\cite{cored, pan2023dfil, GDpreserving, tian2024dynamic, CDDB, hdp}.
Existing methods such as CoReD~\cite{cored}, DFIL~\cite{pan2023dfil}, and SURLID~\cite{surlid} preserve prior knowledge via distillation or feature alignment, but are designed under image-level assumptions.
Unlike deepfake images that contain only spatial artifacts, video deepfakes present complementary evidence along both spatial and temporal axes, whose divergent characteristics make a single entangled representation insufficient to retain during continual updates.
SARB-DF~\cite{prathibha2024sarb}, the only prior effort toward continual deepfake video detection, addresses only spatial cues and follows generic regularization (\eg, EWC~\cite{ewc}) rather than deepfake-specific mechanisms.
These limitations motivate our design that separately preserves spatial, temporal, and spatiotemporal forensic cues throughout continual model updates.

\noindent{\textbf{Continual Learning for Video Representations.}}
Continual learning has been widely studied for video understanding, where models are sequentially adapted to evolving spatial and temporal dynamics~\cite{vclimb, HCE, lee2025essential, stsp, drift}.
These methods are typically designed to preserve high-level semantic representations (\eg, playing tennis) under sequential updates.
However, deepfake video detection depends on modality-sensitive, low-level forensic signals (\eg, frequency artifacts and fine-grained temporal inconsistencies) that are fragile and can be easily washed out under semantics-focused objectives.
As a result, directly applying conventional video continual learning techniques leads to suboptimal adaptation, motivating forensic-aware preservation targets for continual deepfake detection.

\section{Proposed Method}
\label{sec:method}

\begin{figure*}[t!]
\centering
\includegraphics[width=\linewidth]{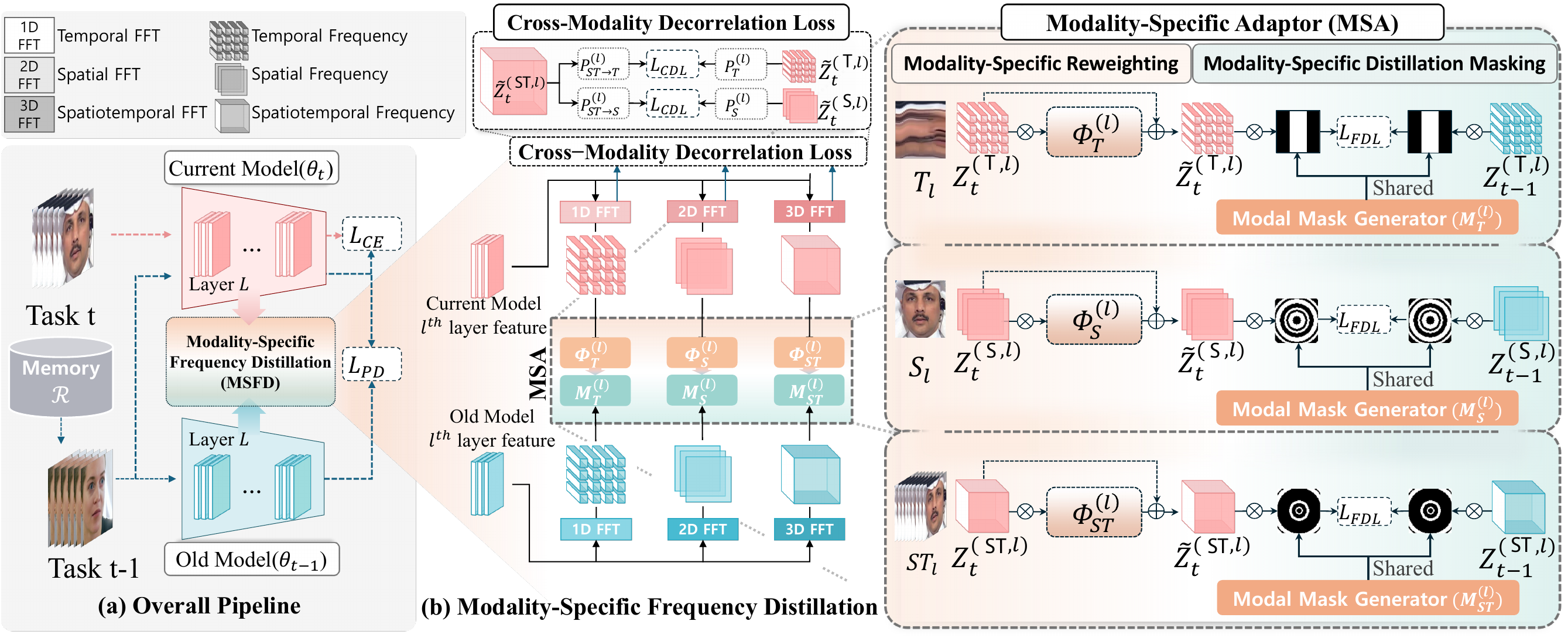}
\caption{\textbf{Our proposed framework, Modality-Specific Frequency Distillation (MSFD) for continual deepfake video detection.} 
(a) Overall pipeline: The current model ($\theta_t$) is trained on task $t$ data and memory ($\mathcal{R}$), distilling knowledge from the frozen previous model ($\theta_{t-1}$). 
(b) Modality-Specific Frequency Distillation (MSFD): Intermediate features from $l^{th}$ layer are decomposed into temporal ($\mathsf{T}$), spatial ($\mathsf{S}$), and spatiotemporal ($\mathsf{ST}$) frequency spectra via 1D, 2D, and 3D FFTs. 
The Modality-Specific Adaptor (MSA) applies modality-specific reweighting ($\Phi_d^{(l)}$) and distillation masking ($M_d^{(l)}$) to compute frequency-domain distillation loss ($\mathcal{L}_{\text{FDL}}$), while the cross-modality decorrelation loss ($\mathcal{L}_{\text{CDL}}$) is computed in parallel. 
}
\label{fig:framework}
\end{figure*}

\label{subsec:problem_formulation}
We address continual deepfake video detection, where the model encounters a sequence of $T$ tasks. At each task $t$, the previous model $\theta_{t-1}$ is frozen as the teacher, and the current model $\theta_t$ is updated using the task data $\mathcal{D}_t$ and a small subset of prior data stored in the replay memory $\mathcal{R}$.
The primary objective at task $t$ is to effectively learn the new deepfake domain introduced in the current dataset, while maintaining high accuracy on all previously learned tasks by preserving knowledge from the previous model $\theta_{t-1}$.

\subsection{Framework Overview}
\label{subsec:overview}
Our proposed framework, Modality-Specific Frequency Distillation (MSFD), employs a unified optimization scheme that leverages frequency-domain distillation and cross-modality regularization. 
The overall pipeline, visualized in Fig.~\ref{fig:framework}, explicitly decomposes intermediate features~\cite{FAM, uhkd} into three complementary modalities---spatial, temporal, and spatiotemporal---via Fast Fourier Transforms (FFT), enabling distillation of distinct deepfake artifacts per modality.
Our framework involves three key mechanisms: (1) the modality-wise frequency transform explicitly decomposes intermediate features into frequency spectra for each modality; (2) the Modality-Specific Adaptor (MSA) performs adaptive calibration and selective masking on the current model's spectra to compute the frequency-domain distillation loss $\mathcal{L}_{FDL}$, thereby more effectively preserving modality-specific cues; (3) the Cross-Modality Decorrelation Loss (CDL) $\mathcal{L}_{CDL}$ promotes distinct spatiotemporal representations while remaining orthogonal to those captured by single-modality branches. These components are integrated with standard classification loss $\mathcal{L}_{CE}$ and logit distillation loss $\mathcal{L}_{PD}$.

\subsection{Modality-wise Frequency Transform}

At task $t$, the current model $\theta_t$ is optimized with respect to the frozen previous model $\theta_{t-1}$ using replay samples from $\mathcal{R}$.
For a set of intermediate layers $L$ in the video backbone, we extract $l^{th}$ layer's feature maps $F_{t-1}^{(l)}, F_t^{(l)} \in \mathbb{R}^{C_l \times T_l \times H_l \times W_l}$ from both models, where $l \in L$, $C_l$ denotes the number of channels, $T_l$ is the temporal dimension, and $H_l, W_l$ are the spatial dimensions at $l^{th}$ layer.

To explicitly decompose these representations into modality-specific components, we apply Fast Fourier Transform (FFT) along different dimensions for feature maps $F_t^{(l)}$ and $F_{t-1}^{(l)}$ to obtain three complementary magnitude spectra for each modality $d \in \{\mathsf{T}, \mathsf{S}, \mathsf{ST}\}$, corresponding to temporal, spatial, and spatiotemporal modalities. We then take the magnitude to construct modality-wise spectra $Z_t^{(d,l)}$ and $Z_{t-1}^{(d,l)}$.

\begin{itemize}
    \item \textbf{Temporal ($\mathsf{T}$):} 1D FFT along the temporal axis $T_l$ to capture temporal frequency patterns.
    \item \textbf{Spatial ($\mathsf{S}$):} 2D FFT over the spatial dimensions $(H_l,W_l)$ to extract spatial frequency characteristics.
    \item \textbf{Spatiotemporal ($\mathsf{ST}$):} 3D FFT across both temporal and spatial axes $(T_l, H_l, W_l)$ to capture joint spatiotemporal frequency, targeting coupled space–time artifacts not recoverable from single-modality spectra alone. 
\end{itemize}

\subsection{Modality-Specific Adaptor}
To more effectively handle modality-specific cues during distillation, we introduce the Modality-Specific Adaptor (MSA), which integrates modality-specific reweighting and modality-specific distillation masking.

\noindent\textbf{Modality-Specific Reweighting.}
For each modality $d \in \{\mathsf{T}, \mathsf{S}, \mathsf{ST}\}$ and $l^{th}$ layer, the modality-specific reweighting module applies a learnable, modality-wise reweighting parameter $\Phi_d^{(l)}$ to recalibrate the current model's spectrum $Z_t^{(d,l)}$ via a residual transformation:
\begin{equation}
\label{eq:msr}
\widetilde{Z}_t^{(d,l)} = \Phi_d^{(l)} \odot Z_t^{(d,l)} + Z_t^{(d,l)},
\end{equation}
where the reweighting parameter $\Phi_d^{(l)}$ is defined with a modality-dependent shape and is automatically broadcast to match the spectrum $Z_t^{(d,l)} \in \mathbb{R}^{B \times C \times T_l \times H_l \times W_l}$, enabling element-wise spectral modulation. 
Specifically, for the temporal modality ($\mathsf{T}$), the reweighting parameter spans only the temporal axis, resulting in $\Phi_\mathsf{T}^{(l)} \in \mathbb{R}^{1 \times C \times T_l \times 1 \times 1}$. For the spatial modality ($\mathsf{S}$), the parameter covers the spatial dimensions, giving $\Phi_\mathsf{S}^{(l)} \in \mathbb{R}^{1 \times C \times 1 \times H_l \times W_l}$. Finally, for the spatiotemporal modality ($\mathsf{ST}$), all axes are reweighted jointly, yielding $\Phi_\mathsf{ST}^{(l)} \in \mathbb{R}^{1 \times C \times T_l \times H_l \times W_l}$.

\noindent\textbf{Modality-Specific Distillation Masking.}
The modality-specific distillation masking module learns a sparse, binary mask $M_d^{(l)}$ to select discriminative frequency bands based on data, rather than hand-crafted priors. The frequency spectrum's radial axis is discretized into $K_d$ bins. 
A learnable logit vector $\mathbf{a}_{d,k}^{(l)} \in \mathbb{R}^2$ for each bin $k$ is converted into a binary pass/block decision $m_{d,k}^{(l)} \in \{0,1\}$ using the Gumbel-Softmax operator $\mathcal{G}(\cdot; \tau)$:
\begin{equation}
\label{eq:mgfm_bin}
m_{d,k}^{(l)} = \mathcal{G}(\mathbf{a}_{d,k}^{(l)}; \tau) \cdot \mathbf{e}_{\text{pass}},
\end{equation}
where $\mathbf{e}_{\text{pass}} = [0,1]^\top$ selects the ``pass'' component, and the temperature $\tau$ is linearly annealed from 1.0 to 0.05 to encourage sharp selections.

The radial mask at frequency location $\mathbf{u}$ is then defined as:
\begin{equation}
\label{eq:msdm}
M_d^{(l)}(\mathbf{u}) = m_{d,\kappa_d^{(l)}(\mathbf{u})}^{(l)} \in \{0,1\},
\end{equation}
which directly maps each frequency location to its corresponding bin's pass/block decision.
The learned radial mask is broadcast to modality-specific dimensions, such as $(1,1,T_l,1,1)$ for temporal, $(1,1,1,H_l,W_l)$ for spatial, and $(1,1,T_l,H_l,W_l)$ for spatiotemporal, and then applied element-wise to each corresponding spectrum to obtain the final masked representations:
\begin{equation}
\widehat{Z}_t^{(d,l)} = M_d^{(l)} \odot \widetilde{Z}_t^{(d,l)}, \quad
\widehat{Z}_{t-1}^{(d,l)} = M_d^{(l)} \odot Z_{t-1}^{(d,l)},
\end{equation}
where $\odot$ denotes element-wise multiplication after broadcasting. We empirically set the radial bins $K_d$ for each modality as $K_{\mathsf{T}}=4$, $K_{\mathsf{S}}=16$, and $K_{\mathsf{ST}}=32$.

\subsection{Training Objective}

\noindent{\textbf{Frequency-domain distillation loss.}}
The frequency-domain distillation loss then enforces knowledge transfer in the most informative spectral regions:
\begin{equation}
\mathcal{L}_{\text{FDL}} = \sum_{l \in \mathcal{L}} \sum_{d \in \{\mathsf{T}, \mathsf{S}, \mathsf{ST}\}} \lambda_d \| \widehat{Z}_t^{(d,l)} - \widehat{Z}_{t-1}^{(d,l)} \|_2^2,
\end{equation}
where $\lambda_d$ is a modality-wise weighting coefficient.

\noindent\textbf{Cross-modality decorrelation loss.}
To ensure the spatiotemporal representation captures unique joint space-time dynamics and remains orthogonal to the single-modality cues, we introduce the Cross-Modality Decorrelation Loss (CDL).
For each $l$-th layer, CDL measures the correlation between the spatiotemporal spectrum and its spatial or temporal counterparts:
\begin{equation}
\label{eq:cdl}
\mathcal{L}_{\mathrm{CDL}} = \sum_{l \in L} \left[
\begin{aligned}
\mathcal{L}_{\mathrm{corr}}(P_{\mathsf{ST} \to \mathsf{S}}^{(l)}(\widetilde{Z}_t^{(\mathsf{ST},l)}), P_{\mathsf{S}}^{(l)}(\widetilde{Z}_t^{(\mathsf{S},l)})) \\
+ \mathcal{L}_{\mathrm{corr}}(P_{\mathsf{ST} \to \mathsf{T}}^{(l)}(\widetilde{Z}_t^{(\mathsf{ST},l)}), P_{\mathsf{T}}^{(l)}(\widetilde{Z}_t^{(\mathsf{T},l)}))
\end{aligned}
\right],
\end{equation}
where $P_d^{(l)}$ represents modality-specific projection heads that map the spectra into an alignment space. 
Before projection, spatial spectra are obtained by averaging over the temporal axis, temporal spectra by averaging across spatial dimensions, and spatiotemporal spectra are used directly without reduction. 
Each spectrum is then projected through its corresponding head ($P_{\mathsf{S}}^{(l)}$, $P_{\mathsf{T}}^{(l)}$, $P_{\mathsf{ST} \to \mathsf{S}}^{(l)}$, $P_{\mathsf{ST} \to \mathsf{T}}^{(l)}$) into an alignment space with dimensionality equal to the number of frequency bins for that modality (\eg., $K_{\mathsf{S}}$, $K_{\mathsf{T}}$).

For projected tensors $A, B \in \mathbb{R}^{B \times C \times k}$, the mean per-channel Pearson correlation is defined as:
\begin{equation}
\mathrm{corr}(A,B) = \frac{1}{BC}\sum_{b=1}^{B}\sum_{c=1}^{C}\left(\frac{1}{k}\sum_{i=1}^{k}A'_{b,c,i}B'_{b,c,i}\right),
\end{equation}
where $A'$ and $B'$ are channel-wise z-score normalized along the frequency dimension.
In practice, $\mathcal{L}_{\mathrm{corr}}(A,B)$ is computed as the mean squared correlation, $\mathcal{L}_{\mathrm{corr}}(A,B) = \mathrm{corr}(A,B)^2$, equally penalizing positive and negative correlations.

\label{subsec:objective}

\noindent{\textbf{Final objective loss.}} The final objective function integrates classification, logit-level distillation~\cite{rebuffi2017icarl, cored}, and our frequency-based regularization terms:
\begin{equation}
\mathcal{L}_{\text{total}} = \mathcal{L}_{\text{CE}} + \lambda_{\text{PD}} \mathcal{L}_{\text{PD}} + \lambda_{\text{FDL}} \mathcal{L}_{\text{FDL}} + \lambda_{\text{CDL}} \mathcal{L}_{\text{CDL}},
\end{equation}
where $\mathcal{L}_{\text{CE}}$ is the standard binary cross-entropy loss computed on the combined data ($\mathcal{D}_t \cup \mathcal{R}$). The term $\mathcal{L}_{\text{PD}}$ is the memory-based logit-level distillation loss, which is applied exclusively to samples from the replay memory $\mathcal{R}$ to align the student's logits with the frozen previous model's predictions. 
Finally, $\lambda_{\text{PD}}$, $\lambda_{\text{FDL}}$, and $\lambda_{\text{CDL}}$ are empirically set weighting coefficients.

\section{Experiments}
\label{sec:experiments}
\noindent\textbf{Datasets.}
We conduct experiments on ten publicly available video-based deepfake datasets spanning a broad range of forgery techniques: FF++~\cite{faceforensics}, DFD~\cite{DFD}, CDF~\cite{CDF}, DFDCp~\cite{DFDC2019Preview}, FFIW~\cite{FFIW}, KoDF~\cite{Kwon_2021_ICCV}, FSh~\cite{faceshifter}, DFo~\cite{dfo}, DF40~\cite{DF40}, and AIGVDBench~\cite{AIGVDBench}.
The first nine are face-centric video datasets capturing the temporal dynamics of facial forgeries; AIGVDBench additionally covers non-facial AI-generated content to evaluate generalization beyond face-centric scenarios.
For DF40~\cite{DF40}, we adopt six manipulation methods: BlendFace~\cite{shiohara2023blendface}, FaceDancer~\cite{facedancer}, and MobileSwap~\cite{mobileswap} for Face Swap (FS), and HyperReenact~\cite{hyperreenact}, MCNet~\cite{mcnet}, and SadTalker~\cite{sadtalker} for Face Reenactment (FR) deepfakes.

\noindent\textbf{Continual Video Deepfake Detection Protocols.}
We evaluate our framework under three complementary continual learning protocols tailored to video-based deepfake detection.
\begin{itemize}
    \item \textit{Protocol 1: Dataset Incremental (DI).}
This protocol emulates a practical scenario where detectors must continually adapt as new deepfake datasets emerge.
The model is sequentially trained on FF++ (Task 1), followed by DFD, CDF, DFDCp, FFIW, and KoDF. For a balanced evaluation, we sample 200 real and 200 fake videos from each newly introduced dataset.

    \item \textit{Protocol 2: Fake Incremental (FI).}
This protocol evaluates robustness against structural shifts across generation paradigms in AIGVDBench~\cite{AIGVDBench}, inspired by~\cite{surlid}: Text-to-Video (T2V) setup using EasyAnimate~\cite{xu2024easyanimate}, Image-to-Video (I2V) via Pyramid-Flow~\cite{pyramidflow}, Video-to-Video (V2V) via LTX~\cite{hacohen2024ltx}, and closed-source generation (SORA~\cite{openai_sora_2024}), each relying on distinct generative priors that produce fundamentally different forgery patterns.
Specifically, subsequent tasks introduce only new fake videos without any real data.

    \item \textit{Protocol 3: Few-shot Incremental (FSI).}
Following the continual deepfake detection setting~\cite{pan2023dfil}, this protocol evaluates adaptation under a constrained data regime (only 25/25 real/fake videos per task, excluding the first task), which mirrors realistic scenarios where new forgeries first appear with limited samples. The model is sequentially updated across FF++, DFDCp, DFD, and CDF.
\end{itemize}

\noindent{\textbf{Evaluation Metrics.}}
For each task \(t\), we compute the video-level AUC, denoted \(AUC_t\), and summarize the overall performance as: $AUC_{all} = \frac{1}{T}\sum_{t=1}^{T} AUC_t.$
Forgetting is measured as the average performance drop relative to when each task was first learned: $F_{avg} = \frac{1}{T-1}\sum_{t=1}^{T-1}\big(a_{t,t} - a_{T,t}\big),$ where $a_{t,t}$ denotes the AUC on Task $t$ immediately after learning it, and \(a_{T,t}\) is the AUC on the same task after completing all \(T\) tasks.  
A higher \(AUC_{all}\) and lower \(F_{avg}\) indicate better retention and a more favorable stability–plasticity balance.
For Protocol 3, following~\cite{pan2023dfil}, we report clip-level accuracy with Average Accuracy (AA) and Average Forgetting (AF), where AA is the mean accuracy over all seen tasks, and AF measures the average performance drop from when each task was first learned.

\begin{table*}[t!]
\centering
\caption{\textbf{Comparison under Protocol~1: Dataset-Incremental (DI).}
We evaluate continual adaptation as new deepfake video datasets are sequentially introduced, and report video-level AUC (\%) after the final task, overall $AUC_{all}$, and average forgetting $F_{avg}$.
Unless otherwise noted, all methods use 3DResNet-18 as backbone; all replay-based methods set memory size to 200 clips.
Best results are in \textbf{bold} and second best are \underline{underlined}. $\uparrow$/$\downarrow$ indicates higher/lower is better.}
\label{tab:di_comparison}
\resizebox{0.88\linewidth}{!}{
\begin{tabular}{@{} l c !{\vrule width 0.6pt}c c c c c c !{\vrule width 0.6pt} c c @{}}
\toprule
\textbf{Method} & \textbf{Backbone} & \textbf{FF++} & \textbf{DFD} & \textbf{CDF} & \textbf{DFDCp} & \textbf{FFIW} & \textbf{KoDF} & $\boldsymbol{AUC_{all}\!\uparrow}$ & $\boldsymbol{F_{avg}\!\downarrow}$ \\
\midrule
\multicolumn{2}{@{}l}{$\vartriangleright$\textit{General continual learning methods}}  \\[1pt]
Finetuning &\multirow{6}{*}{3DResNet-18} & 89.47 & 89.29 & 71.13 & 73.82 & 84.98 & 99.77 & 84.74 & 14.07 \\
EWC~\cite{ewc} & & 81.76 & 86.16 & 67.67 & 69.07 & 83.23 & \underline{99.89} & 81.30 & 18.69 \\
ER~\cite{er} & & 92.36 & 94.21 & 72.14 & 74.13 & 91.68 & 99.75 & 87.38 & 10.77 \\
iCaRL-NME~\cite{rebuffi2017icarl} & & 86.67 & 92.22 & 74.76 & 80.59 & 87.60 & 96.86 & 86.45 & 8.14 \\
iCaRL-FC~\cite{rebuffi2017icarl} & & 96.82 & 94.65 & 80.50 & 79.20 & 88.77 & 99.07 & 89.84 & 6.81 \\
DCE~\cite{li2025addressing}& & \textbf{99.63}& 96.05 & 72.97 & 83.93 & 90.40 & 90.91 & 88.98 & \textbf{0.07} \\
IDER~\cite{liu2026ider} & & 95.32 & \textbf{96.26}& \textbf{93.62} & 80.09 & 83.80 & 97.48 & \underline{91.10} & 2.97 \\
\midrule
\multicolumn{2}{@{}l}{$\vartriangleright$\textit{Video continual learning methods}}   \\[1pt]
vCLIMB~\cite{vclimb} &\multirow{3}{*}{3DResNet-18} & 80.07 & 90.46 & 66.61 & 77.71 & 85.51 & 96.36 & 82.79 & 11.27 \\
Drift~\cite{drift} & & 85.03 & 82.80 & 67.56 & 76.14 & \underline{94.36} & \textbf{99.99} & 84.31 & 14.53 \\
HCE~\cite{HCE} & & 81.54 & 91.56 & 62.18 & 78.96 & 92.08 & 99.28 & 84.27 & 13.06 \\
STSP~\cite{stsp} & & \underline{99.48} & 92.67 & 83.76 & 77.15 & 82.78 & 98.35 & 89.03 & 3.98 \\\hdashline
ESSENTIAL~\cite{lee2025essential} & ViT-B/16 (CLIP) & 68.10 & \underline{95.50} & 74.40 & \textbf{89.70} & 41.30 & 99.30 & 78.05 & 20.04 \\
\midrule
\multicolumn{2}{@{}l}{$\vartriangleright$\textit{Continual deepfake detection methods}}    \\[1pt]
DFIL~\cite{pan2023dfil} &\multirow{3}{*}{3DResNet-18}& 90.51 & 92.02 & 78.86 & 74.58 & 91.52 & 99.76 & 87.88 & 10.02 \\
HDP~\cite{hdp} & & 92.22 & 86.20 & 68.58 & 70.21 & 83.07 & 99.95 & 83.37 & 16.14 \\
SURLID~\cite{surlid} & & 94.98 & 90.25 & 83.20 & 81.30 & 76.81 & 99.12 & 87.61 & 8.19 \\\hdashline
SARB-DF~\cite{prathibha2024sarb} & Xception + Transformer & 81.99 & 90.60 & 83.26 & 83.55 & 80.46 & 99.16 & 86.50 & 11.63 \\
\midrule
\textbf{Ours} & 3DResNet-18 & 98.42 & 91.15 & \underline{90.17} & \underline{84.14} & \textbf{94.88} & 99.36 & \textbf{93.02} & \underline{2.29} \\
\bottomrule
\end{tabular}
}
\end{table*}

\noindent{\textbf{Comparison Methods.}}
We compare against a broad spectrum of continual learning baselines: naive fine-tuning, EWC~\cite{ewc}, DCE~\cite{li2025addressing}, replay-based methods (ER~\cite{er}, iCaRL-NME/FC~\cite{rebuffi2017icarl}, IDER~\cite{liu2026ider}), video continual learning methods (Drift~\cite{drift}, HCE~\cite{HCE}, vCLIMB~\cite{vclimb}, STSP~\cite{stsp}, ESSENTIAL~\cite{lee2025essential}), and continual deepfake detection methods (SARB-DF~\cite{prathibha2024sarb}, DFIL~\cite{pan2023dfil}, HDP~\cite{hdp}, SURLID~\cite{surlid}).
Unless otherwise specified, all methods adopt 3D ResNet-18~\cite{3dresnet} and are initialized from Task~1 weights for a fair comparison.
SARB-DF and ESSENTIAL are evaluated with their native backbones, as their architectures cannot be directly adapted.

\noindent{\textbf{Implementation Details.}}
For preprocessing, as described in~\cite{ftcn}, we detect and align faces using RetinaFace. The model processes a 32-frame clip at a spatial resolution of \(224 \times 224\). During training, up to 270 frames are extracted from each video to form multiple 32-frame clips via temporal sampling, while evaluation uses up to 110 frames to generate non-overlapping or minimally overlapping clips. 
Video-level scores are computed by averaging the clip-level logits. 
All models are trained for 20 epochs per task using the Adam optimizer with a learning rate of \(1\times10^{-4}\) and a batch size of 8. 
The replay memory stores 200 clips (each containing 32 frames) for replay-based continual learning methods.
We adopt iCaRL~\cite{rebuffi2017icarl} with a linear classifier as a baseline and employ its exemplar-based memory update strategy to focus on the effectiveness of our preservation mechanism.
Additional implementation details are provided in the Supplementary.

\subsection{Comparisons on Continual Deepfake Video Detection}

\noindent\textbf{Comparison on Protocol 1: Dataset Incremental (DI).}
As shown in Table~\ref{tab:di_comparison}, our method consistently outperforms existing continual learning strategies under Protocol 1, achieving the highest $AUC_{all}$.
Conventional video continual learning methods (\eg, vCLIMB~\cite{vclimb}, HCE~\cite{HCE}) primarily focus on preserving high-level semantic representations, which are insufficient for retaining low-level forensic cues, resulting in pronounced forgetting.
Continual deepfake detection methods (\eg, DFIL~\cite{pan2023dfil}, SURLID~\cite{surlid}) rely on a single joint representation under image-level assumptions, failing to disentangle spatial and temporal forensic signals.
MSFD overcomes these limitations by independently preserving modality-specific forensic cues in the frequency domain, achieving the highest overall $AUC_{all}$.
While DCE~\cite{li2025addressing} and IDER~\cite{liu2026ider} show competitive forgetting, this comes at the cost of reduced plasticity.
Furthermore, as shown in Fig.~\ref{fig:performanceplot}, our method maintains near-zero forgetting throughout the sequence while achieving a steady increase in AUC, demonstrating both stability and adaptability.

\begin{table}[t!]
\centering
\begin{minipage}[t]{0.48\columnwidth}
\centering
\caption{\textbf{Protocol 2: Fake-Incremental.}
We evaluate continual adaptation to structurally distinct generation paradigms and report AUC (\%) after the final task, overall $AUC_{all}$, and average forgetting $F_{avg}$.}
\label{tab:protocol2}
\resizebox{\linewidth}{!}{
\begin{tabular}{l|cccc|cc}
\toprule
\textbf{Method} & \textbf{T2V} & \textbf{I2V} & \textbf{V2V} & \textbf{SORA} & $\boldsymbol{AUC_{all}\!\uparrow}$ & $\boldsymbol{F_{avg}\!\downarrow}$ \\
\midrule
iCaRL-FC~\cite{rebuffi2017icarl} & 98.49 & \underline{87.99}& 96.51 & 96.71 & \underline{94.93}& 0.30 \\
IDER~\cite{liu2026ider} & \textbf{99.69} & 84.87& 95.92 & 96.51 & 94.25& \underline{-0.32} \\
STSP~\cite{stsp}                 & \underline{99.57} & 83.28 & 93.65 & 89.53 & 91.51 & \textbf{-0.37} \\ 
DFIL~\cite{pan2023dfil}          & 93.24 & 83.88 & 95.35 & \underline{97.74}& 92.55 & 4.59 \\
SURLID~\cite{surlid}             & 97.31 & 83.94 & \underline{96.97}& 95.62 & 93.46 & 0.25\\
\midrule
\textbf{Ours}                    & 98.89& \textbf{91.47} & \textbf{97.49} & \textbf{98.84} & \textbf{96.67} & 1.37 \\
\bottomrule
\end{tabular}
}
\label{tab:protocol2_fake_incremental}
\end{minipage}
\hfill
\begin{minipage}[t]{0.48\columnwidth}
\centering
\caption{\textbf{Protocol 3: Few-shot Incremental.}
We evaluate continual adaptation under a restricted data regime and report clip-level accuracy (\%) after the final task, Average Accuracy (AA), and Average Forgetting (AF), following~\cite{pan2023dfil}.}
\label{tab:protocol3}
\resizebox{\linewidth}{!}{
\begin{tabular}{l|cccc|cc}
\toprule
\textbf{Method} & \textbf{FF++} & \textbf{DFDCp} & \textbf{DFD} & \textbf{CDF} & \textbf{AA} $\uparrow$ & \textbf{AF} $\downarrow$ \\
\midrule
iCaRL-FC~\cite{rebuffi2017icarl} & 95.95 & 68.23 & 88.08 & 75.53& 81.95 & 2.39 \\
IDER~\cite{liu2026ider} & 95.50 & \textbf{77.04} & \textbf{94.67} & 66.67 & \underline{83.47} & \textbf{-2.10} \\
STSP~\cite{stsp}                 & 87.81 & \underline{76.78} & 52.86 & \textbf{82.90} & 75.09 & 1.59 \\
DFIL~\cite{pan2023dfil}          & 94.50 & 63.14 & 81.93 & \underline{79.08}& 79.66 & 5.47 \\
SURLID~\cite{surlid}             & \underline{95.98}& 71.91 & \underline{93.19} & 71.24 & 83.08 & 0.58\\
\midrule
\textbf{Ours}                    & \textbf{97.10} & 72.04 & 90.32 & 75.13 & \textbf{83.65} & \underline{-1.25} \\
\bottomrule
\end{tabular}
}
\label{tab:protocol3_fewshot}
\end{minipage}
\end{table}

\noindent\textbf{Comparison on Protocol 2: Fake Incremental (FI).}
MSFD achieves the highest overall $AUC_{all}$ across generation paradigms (Table~\ref{tab:protocol2}).
We select the five most competitive methods from Protocol 1, including iCaRL-FC~\cite{rebuffi2017icarl}, IDER~\cite{liu2026ider}, STSP~\cite{stsp}, DFIL~\cite{pan2023dfil}, and SURLID~\cite{surlid}, for evaluation under Protocols 2 and 3.
The gain is particularly notable on I2V, where MSFD outperforms all baselines by a large margin, suggesting that modality-specific frequency distillation effectively preserves generation-paradigm-specific forensic cues that single-representation approaches fail to retain across tasks.
While STSP and IDER report lower forgetting, this comes at the cost of substantially lower overall accuracy, indicating that they sacrifice adaptability to new paradigms to retain prior knowledge.
MSFD achieves a more favorable balance, attaining the highest $AUC_{all}$ with moderate forgetting.

\noindent\textbf{Comparison on Protocol 3: Few-shot Incremental (FSI).}
Under the few-shot setting (Table~\ref{tab:protocol3}), MSFD achieves the highest AA while also exhibiting negative forgetting (AF = $-$1.25), indicating positive backward transfer to previously learned tasks.
Although IDER reports a lower AF, its overall AA is lower than that of MSFD, suggesting that MSFD provides a stronger few-shot adaptability.
Notably, STSP, which showed competitive performance in Protocol 1, suffers significant degradation under both Fake-Incremental and Few-shot Incremental settings, revealing that conventional video continual learning approaches lack the robustness to handle the extreme distribution shifts and data scarcity inherent in evolving deepfake scenarios.

\subsection{Ablation Study and Analysis}
\label{subsec:ablation}
\begin{figure}[t!]
  \centering
  \begin{minipage}[t]{0.48\columnwidth}
    \centering
    \includegraphics[width=\linewidth]{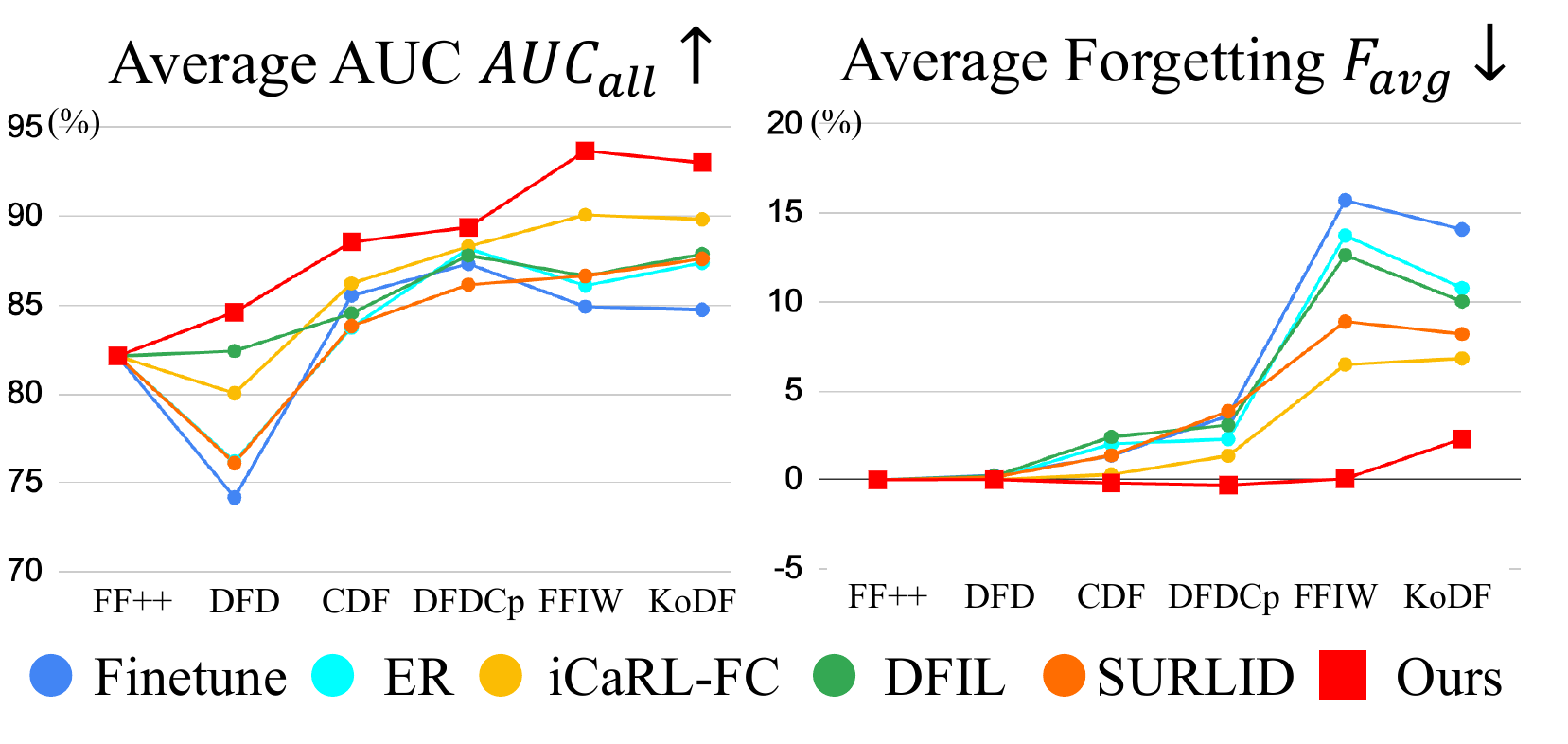}
    \caption{\textbf{Performance trends across sequential tasks.}
    We visualize the average performance $AUC_{all}$ and forgetting $F_{avg}$  across sequential tasks under Protocol~1.}
    \label{fig:performanceplot}
  \end{minipage}
  \hfill
  \begin{minipage}[t]{0.48\columnwidth}
    \centering
    \includegraphics[width=\linewidth]{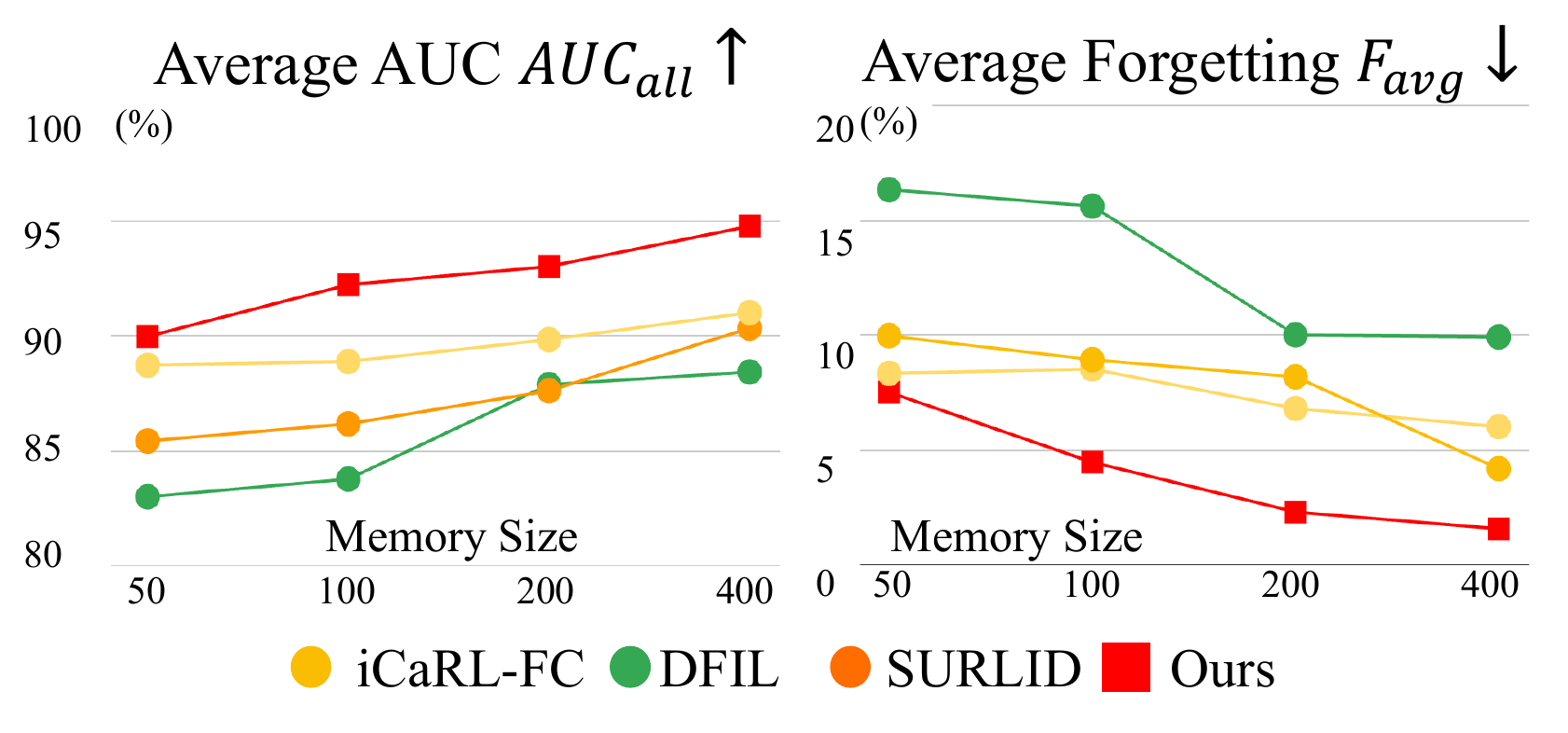}
    \caption{\textbf{Sensitivity for memory buffer size.}
    We report average performance $AUC_{all}$ and forgetting $F_{avg}$ across a memory size from 50 to 400 in Protocol~1.
    }
    \label{fig:memory_size}
  \end{minipage}
\end{figure}

We conduct ablation studies to validate the contribution of each component in our framework. 
Further analyses are provided in the supplementary material.

\noindent\textbf{Analysis on Memory Size Sensitivity.}
We analyze the sensitivity of our framework to the size of the replay memory buffer. As shown in Fig.~\ref{fig:memory_size}, our method (red) consistently outperforms all baselines across memory sizes from 50 to 400 in both $AUC_{all}$ and $F_{avg}$ under Protocol 1.
Most notably, with only 50 memory samples, our framework already matches or surpasses DFIL and SURLID operating with 400 samples. 
This efficiency stems from our modality-specific distillation, which focuses on the most discriminative spectral components as replay exemplars, preserving critical knowledge without requiring large memory budgets.
These results confirm that MSFD remains effective even in resource-constrained continual learning scenarios.

\begin{wrapfigure}{r!}{0.45\columnwidth}
  \centering
\includegraphics[width=0.45\columnwidth]{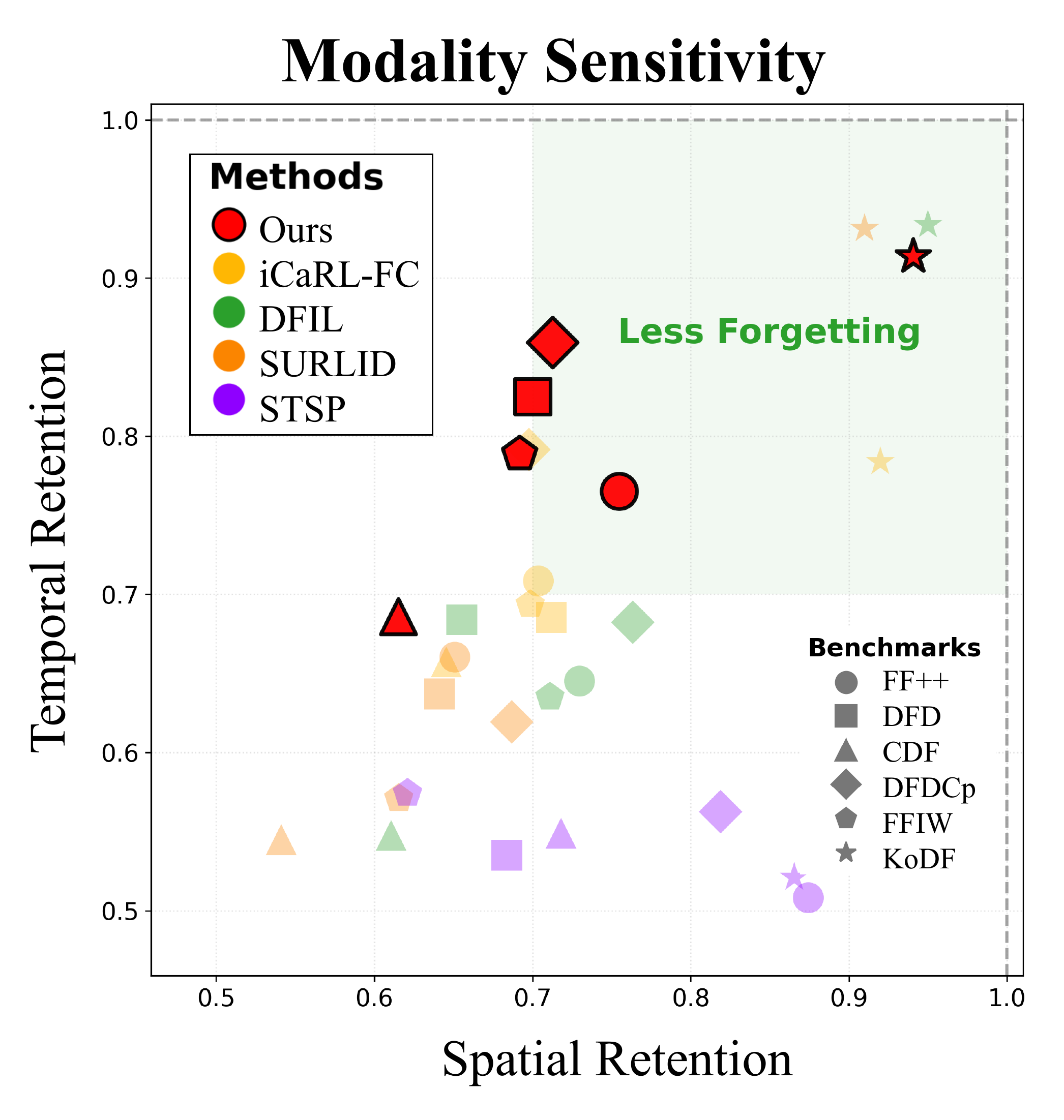}
  \caption{\textbf{Forgetting analysis across spatial and temporal modalities.} Each point represents a method–dataset pair, plotting spatial and temporal retention, defined as the ratio of the final model's AUC under each perturbation to its AUC when first learned.}
    \label{fig:perturrobust}
\end{wrapfigure}
\noindent\textbf{Forgetting Analysis Across Modalities.}
As shown in Fig.~\ref{fig:perturrobust}, our method consistently occupies the upper-right across most benchmarks, demonstrating superior preservation of spatial and temporal forgery cues under continual updates.
To analyze modality-specific forgetting behavior beyond aggregate metrics, Fig.~\ref{fig:perturrobust} visualizes the spatial and temporal retention of each method across benchmarks in Protocol 1, defined as the ratio of the final model's AUC under spatial and temporal perturbation (grid shuffling/frame shuffling) to its AUC when first learning each dataset.
While competing methods scatter toward lower retention on one or both modalities, our results confirm that explicitly preserving spatial and temporal representations is critical for robust continual deepfake detection. Details will be provided in the supplementary material.

\begin{table*}[t]
\centering
\caption{\textbf{Ablation studies of MSFD.}
(a) Contribution of major components (MSFD, MSA, and CDL) over the baseline.
Multi-layer Distillation extends naive feature distillation to multiple intermediate layers of the baseline.
(b) Effect of modality decomposition: $\mathsf{S}$ (spatial), $\mathsf{T}$ (temporal), and $\mathsf{ST}$ (spatiotemporal).
(c) Effect of CDL on Protocol~1 and robustness on FF++/DF40 (FS: face swap, FR: face reenactment).
(d) Breakdown of MSA (MSR and MSDM).
(e) Mask types: fixed high/low-pass vs.\ learnable Gumbel--Softmax in MSDM.
(f) Backbone generalization under continual learning settings.
Unless otherwise specified, all results are reported under Protocol~1.
}
\label{tab:ablation_all}
\begin{minipage}[t]{0.36\linewidth}
\centering
(a) Components.\\[1pt]
\label{tab:ablation_component}
\resizebox{\linewidth}{!}{%
\begin{tabular}{l|ccc|cc}
\toprule
Model & MSFD & MSA & CDL & $AUC_{all}\!\uparrow$ & $F_{avg}\!\downarrow$ \\
\midrule
Baseline  &  &  &  & 89.84 & 6.81 \\
Multi-layer Distill. &  &  &  & 90.64 & 6.34 \\
MSFD (Naive) & \checkmark &  &  & 92.65 & 4.09 \\
MSFD + MSA & \checkmark & \checkmark &  & 92.14 & 3.20 \\
MSFD + CDL & \checkmark &  & \checkmark & 92.77 & 2.65 \\
\midrule
\textbf{Ours (Full)} & \checkmark & \checkmark & \checkmark & \textbf{93.02} & \textbf{2.29} \\
\bottomrule
\end{tabular}}
\end{minipage}
\hfill
\begin{minipage}[t]{0.30\linewidth}
\centering
(b) Modalities.\\[1pt]
\label{tab:abaltion_modality}
\resizebox{\linewidth}{!}{%
\begin{tabular}{l|ccc|cc}
\toprule
Modality & $\mathsf{S}$ & $\mathsf{T}$ & $\mathsf{ST}$ & $AUC_{all}\!\uparrow$ & $F_{avg}\!\downarrow$ \\
\midrule
w/o Decompose & --  & -- & -- & 90.64 & 6.34 \\ 
Spatial Only & \checkmark & -- & -- & 90.69 & 5.29 \\
Temporal Only & -- & \checkmark & -- & 91.45 & 4.41 \\
Spatiotemp. Only & -- & -- & \checkmark & 92.06 & 3.27 \\
\midrule
\textbf{Full (Ours)} & \checkmark & \checkmark & \checkmark & \textbf{93.02} & \textbf{2.29} \\
\bottomrule
\end{tabular}}
\end{minipage}
\hfill
\begin{minipage}[t]{0.30\linewidth}
\centering
(c) Effect of CDL.\\[1pt]
\label{tab:abaltion_cdl}
\resizebox{\linewidth}{!}{%
\begin{tabular}{l|cc|cc|cc}
\toprule
\multirow{2}{*}{Method}
& \multicolumn{2}{c}{Protocol 1}
& \multicolumn{2}{c}{FF++}
& \multicolumn{2}{c}{DF40} \\
\cmidrule(lr){2-3}\cmidrule(lr){4-5}\cmidrule(lr){6-7}
 & $AUC_{all}\!\uparrow$ & $F_{avg}\!\downarrow$ & FS & FR & FS & FR \\ \midrule
w/o CDL & 92.14 & 3.20 & 99.5 & 97.5 & 92.8 & 75.6 \\
\textbf{Ours} & \textbf{93.02} & \textbf{2.29} & \textbf{99.6} & \textbf{97.8} & \textbf{93.3} & \textbf{85.6} \\
\bottomrule
\end{tabular}}
\end{minipage}

\begin{minipage}[t]{0.33\linewidth}
\centering
(d) MSA components.\\[2pt]
\resizebox{\linewidth}{!}{%
\begin{tabular}{l|cc|cc}
\toprule
Configuration & MSR & MSDM & $AUC_{all}\!\uparrow$ & $F_{avg}\!\downarrow$ \\
\midrule
w/o MSA & -- & -- & 92.77 & 2.65 \\
+ MSR Only & \checkmark & -- & 92.97 & 2.43 \\
+ MSDM Only & -- & \checkmark & 91.91 & 3.08 \\
\midrule
\textbf{MSA (Full)} & \checkmark & \checkmark & \textbf{93.02} & \textbf{2.29} \\
\bottomrule
\end{tabular}}
\end{minipage}
\hfill
\begin{minipage}[t]{0.3\linewidth}
\centering
(e) Mask type in MSDM.\\[2pt]
\resizebox{\linewidth}{!}{%
\begin{tabular}{l|c|cc}
\toprule
Mask Type & Trainable & $AUC_{all}\!\uparrow$ & $F_{avg}\!\downarrow$ \\
\midrule
None & -- & 92.97 & 2.43 \\
High-pass & Fixed & 92.94 & 2.53 \\
Low-pass & Fixed & 92.38 & 3.14 \\
\midrule
\textbf{Ours} & \textbf{Learnable} & \textbf{93.02} & \textbf{2.29} \\
\bottomrule
\end{tabular}}
\end{minipage}
\hfill
\begin{minipage}[t]{0.3\linewidth}
\centering
(f) Backbone.\\[2pt]
\resizebox{\linewidth}{!}{%
\begin{tabular}{l|l|cc}
\toprule
Backbone & Method & $AUC_{all}\!\uparrow$ & $F_{avg}\!\downarrow$ \\
\midrule
\multirow{2}{*}{R3D-18} & iCaRL-FC & 89.84 & 6.81 \\
 & Ours & \textbf{93.02} & \textbf{2.29} \\
\midrule
\multirow{2}{*}{R(2+1)D-18} & iCaRL-FC & 86.10 & 8.75 \\
 & Ours & \textbf{89.24} & \textbf{2.17} \\
\midrule
\multirow{2}{*}{FTCN} & iCaRL-FC & 74.05 & 16.39 \\
 & Ours & \textbf{75.34} & \textbf{8.03} \\
\bottomrule
\end{tabular}}
\end{minipage}
\end{table*}

\noindent\textbf{Contribution of Components.}
Each proposed component addresses a distinct bottleneck, and their combination achieves the best balance between adaptation and retention (Table~\ref{tab:ablation_all}-(a)). Multi-layer feature distillation provides only marginal improvement, confirming that preserving entangled spatial-temporal features in the raw space is insufficient. MSFD (Naive) yields a substantial gain by decomposing features into modality-specific frequency spectra, validating frequency-domain decomposition as the key enabler. MSA improves stability by suppressing noisy frequency bands, while CDL reduces cross-modality redundancy. The full model achieves the strongest overall performance, confirming their complementary roles.

\noindent\textbf{Analysis of Modality Decomposition.}
All three frequency modalities are necessary, and their combination yields the most robust performance (Table~\ref{tab:ablation_all}-(b)).
Spatial-only ($\mathsf{S}$) provides the smallest gain, as spatial cues are well preserved by standard distillation. 
Temporal-only ($\mathsf{T}$) shows a larger improvement, consistent with prior findings that temporal cues are more generalizable forensic signals~\cite{ftcn, lips}. 
Spatiotemporal-only ($\mathsf{ST}$) achieves the strongest result by capturing joint space-time dynamics inaccessible to 1D or 2D transforms. 
The full model further improves, confirming complementarity across the three spectra.

\noindent\textbf{Effect of Cross-modality Decorrelation Loss (CDL).}
CDL improves both accuracy and stability by preventing the $\mathsf{ST}$ branch from redundantly encoding cues already preserved by the spatial and temporal branches (Table~\ref{tab:ablation_all}-(c)). 
Without CDL, such redundancy wastes capacity and compounds forgetting when overlapping representations shift during later tasks. 
Furthermore, we find the largest improvement on Face Reenactment (FR) in DF40~\cite{DF40}, an advanced variant of FF++. 
Since FR modifies expression dynamics, its artifacts appear as temporally inconsistent spatial deformations, making spatiotemporal cues more critical than in Face Swap (FS).

\noindent\textbf{Analysis of the Modality-Specific Adaptor (MSA).}
The combination of Modality-Specific Reweighting (MSR) and Modality-Specific Distillation Masking (MSDM) achieves the highest performance and lowest forgetting, demonstrating strong synergy between the two components as shown in Table~\ref{tab:ablation_all}-(d).
MSR adaptively calibrates channel-wise spectral responses to reduce forgetting through modality-aware spectral reweighting, while MSDM selectively highlights discriminative frequency components via a Gumbel–Softmax mechanism. 
Their complementary roles, MSR for calibration and MSDM for frequency-wise selection, enable comprehensive spectral adaptation across all modalities.

To further validate the learnable masking mechanism, we compare the modality-specific distillation masking against a fixed frequency filter. 
As shown in Table~\ref{tab:ablation_all}-(e), replacing the learnable mask with a static high-pass or low-pass filter degrades performance, indicating that fixed frequency selection cannot adapt to task-specific spectral characteristics. In contrast, our learnable mask dynamically allocates attention across frequency regions, achieving the best performance and confirming that adaptive frequency selection is crucial for capturing discriminative spectral cues and maintaining stability in continual deepfake video detection.

\begin{figure}[t!]
  \centering
\includegraphics[width=\columnwidth]{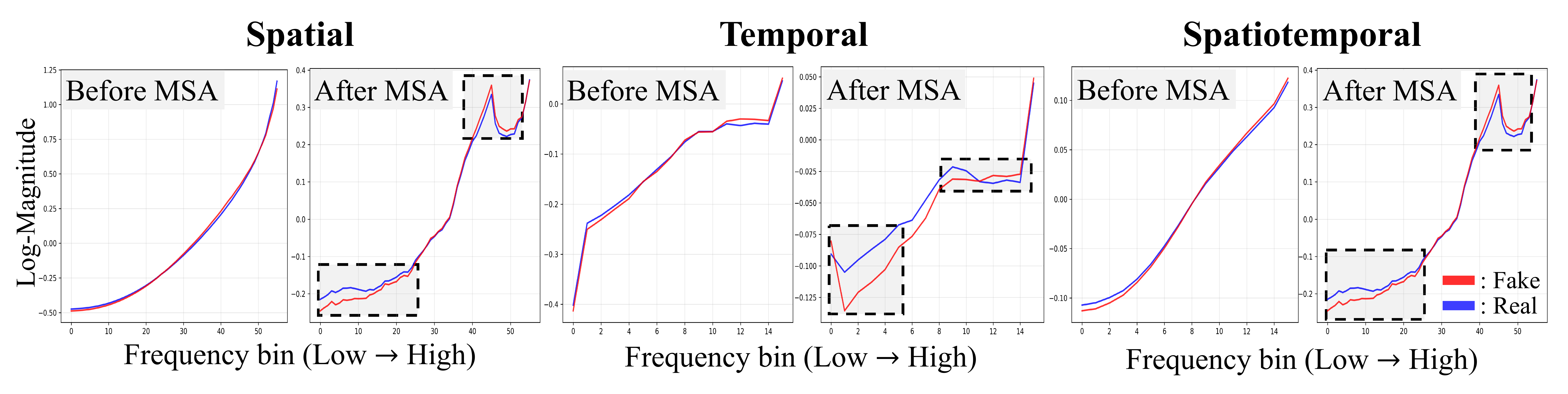}
    \caption{\textbf{Effect of the Modality-Specific Adaptor (MSA) on frequency distributions.}
We visualize frequency spectra of real (blue) and fake (red) samples at Layer~1 across three modalities (spatial, temporal, and spatiotemporal).
For each modality, the left plot shows spectra before MSA and the right plot after MSA.
}
\label{fig:frequency}
\end{figure}

To analyze the impact of MSA, we visualize the frequency distributions of real and fake samples at Layer~1, representing low-level features, across spatial, temporal, and spatiotemporal modalities using our model. As shown in Fig.~\ref{fig:frequency}, the top row presents the distributions before applying MSA, while the bottom row illustrates those after applying MSA. Without MSA, the spectral responses of real and fake samples largely overlap. In contrast, after MSA is introduced, the separation between real and fake distributions becomes more pronounced across all modalities, demonstrating that MSA enables the model to learn forgery-relevant cues even from the low-level representation stage.

\noindent\textbf{Backbone Generalization Analysis.}
Our modality-aware frequency distillation generalizes across architecturally diverse backbones, consistently outperforming the baseline on all tested networks (Table~\ref{tab:ablation_all}-(f)).
To further validate this generality, we evaluate our framework across different backbone architectures under identical continual learning settings, including the standard 3D ResNet-18 and R(2+1)D-18 video backbones~\cite{3dresnet}, as well as the FTCN~\cite{ftcn}, a deepfake-specific backbone. Note that FTCN is originally equipped with a transformer, from which we remove the transformer modules due to memory constraints.

\noindent{\textbf{Failure Case.}}
Despite consistent overall gains, our proposed method exhibits two failure modes. 
\emph{(i) Race-specific distribution shifts}: after learning KoDF (predominantly Asian faces), MSFD's forgetting rises from near-zero to $\sim$2 (Fig.~\ref{fig:performanceplot}).
This suggests a demographic shift in KoDF: differences in facial appearance change the spatial/frequency patterns learned earlier, causing more forgetting.
\emph{(ii) Residual asymmetric retention}: the spatial and temporal branches still degrade at uneven rates (Fig.~\ref{fig:perturrobust}).

\section{Conclusion and Future Work}
\label{sec:conclusion}

In this work, we introduced a novel framework for continual deepfake video detection that mitigates catastrophic forgetting through modality-specific frequency-domain knowledge preservation.
Specifically, we proposed a Modality-Specific Frequency Distillation (MSFD) strategy that decomposes intermediate representations into spatial, temporal, and spatiotemporal spectra, enabling modality-specific knowledge transfer.
We further introduced the Modality-Specific Adaptor (MSA) for discriminative spectral calibration and masking, along with the Cross-Modality Decorrelation Loss (CDL) to encourage orthogonal frequency representations across modalities.
Through comprehensive experiments across multiple continual learning protocols, unseen forgery domains, and perturbation conditions, our method demonstrated superior generalization, robustness, and stability compared to existing continual deepfake detection frameworks.

\noindent{\textbf{Future Work.}}
Building upon the strong performance of memory-based replay methods in continual deepfake video detection, future work will explore deepfake video-specific memory update strategies that explicitly consider both spatial and temporal information to guide exemplar selection and maintenance.
Additionally, developing adaptive mechanisms that dynamically balance the importance of each modality per task could further improve robustness across diverse generation paradigms and reduce sensitivity to hyperparameter choices.

\section*{Acknowledgements}
This work was supported in part by the Institute of Information \& Communication Technology Planning \& Evaluation (IITP) grant funded by the Korea government (MSIT) (No. RS-2025-02263841, Development of a Real-time Multimodal Framework for Comprehensive Deepfake Detection Incorporating Common Sense Error Analysis; RS-2021-II211341, Artificial Intelligence Graduate School Program (Chung-Ang University)), and by [Ministry of Science and ICT] (MSIT) through the [National Research Foundation of Korea] (NRF) funded by the Ministry of Science and ICT (RS-2025-16066667).

\clearpage
\begin{center}
\centering 
\Large
\textbf{Preserving Knowledge across Space and Time \\for Continual Video Deepfake Detection} \\
\vspace{0.5em}Supplementary Material \\
\vspace{1.0em}
\end{center}
\setcounter{page}{1}
\setcounter{figure}{0}
\setcounter{equation}{0}
\setcounter{table}{0}
\setcounter{section}{0}
\renewcommand{\thefigure}{\Alph{figure}}
\renewcommand{\thetable}{\Alph{table}}
\renewcommand{\theequation}{\arabic{equation}s}
\renewcommand{\thesection}{\Alph{section}}

\section{Additional Experiments}

\subsection{Additional Analysis}

\subsubsection{Visualization of Representation Preservation}
\begin{wrapfigure}{r!}{0.48\columnwidth}
  \centering
\vspace{-9mm}
    \includegraphics[width=\linewidth]{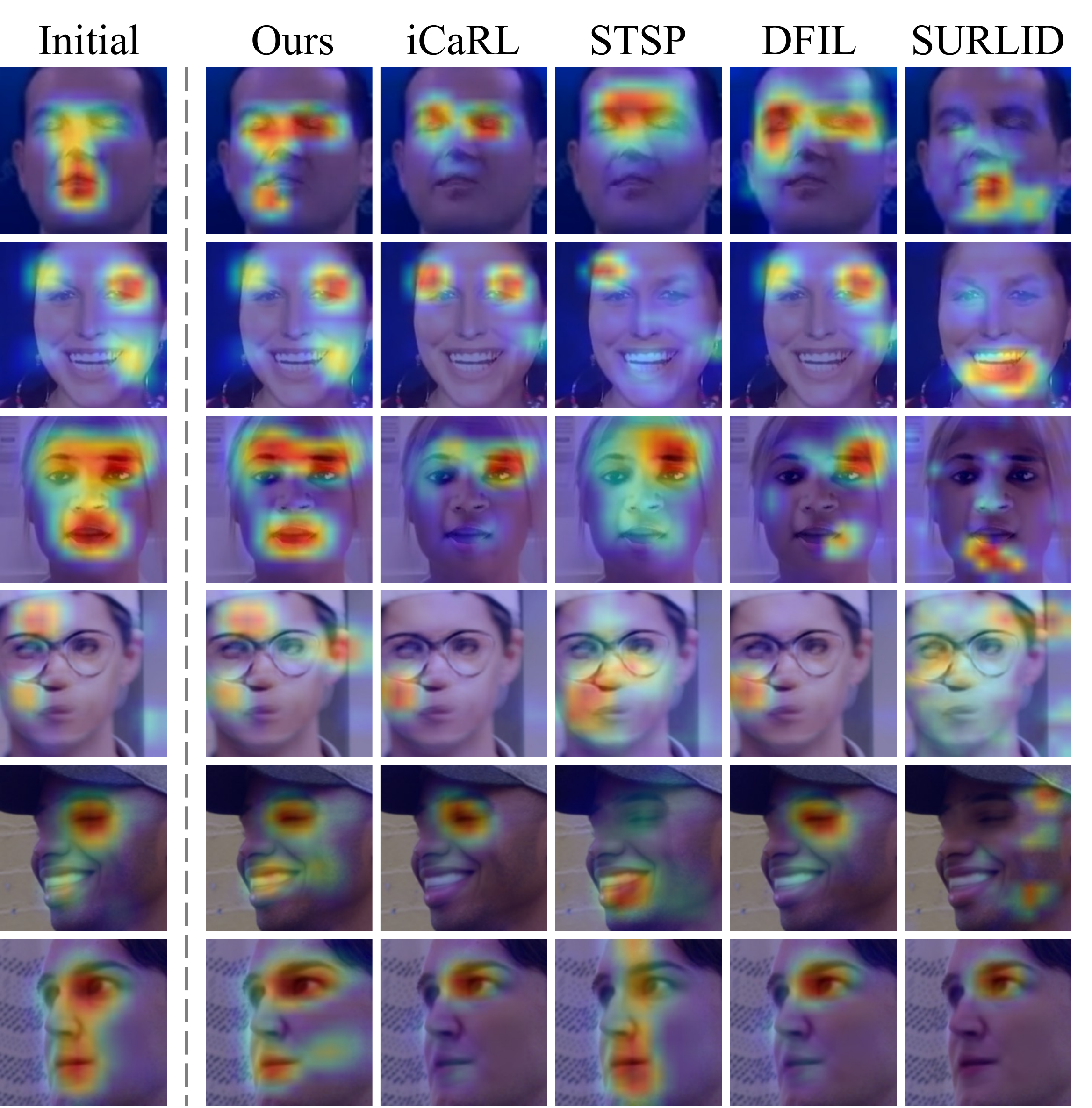}
  \caption{\textbf{Grad-CAM visualization of representation preservation}. 
    We compare Grad-CAM attention maps on FF++ (Task 1) between the initial model and each continual learning method after completing all tasks.
    }
  \label{fig:gradcam}
  \vspace{-9mm}
\end{wrapfigure}

To examine how each method retains previously learned knowledge, we visualize Grad-CAM~\cite{gradcam} activation maps from the initial task (FF++) before and after continual training (Fig.~\ref{fig:gradcam}).
While existing methods exhibit significant drift in attention, shifting focus to irrelevant regions or losing sensitivity to deepfake artifacts, our approach maintains consistent activation around key forgery regions such as the mouth, eyes, and facial boundaries.
This qualitative result highlights the effectiveness of our modality-specific frequency distillation in preserving early spatial–temporal representations even after multiple continual updates.

\subsubsection{Generalization for Unseen Datasets}
As shown in Table~\ref{tab:unseen_generalization}, our method achieves the highest average AUC across all unseen forgery datasets, outperforming prior continual detectors such as DFIL~\cite{pan2023dfil} and SURLID~\cite{surlid}. 
To assess unseen deepfake generalization, models that performed highly under Protocol 1 were trained under Protocol 1 up to the DFDCp stage and subsequently evaluated on four unseen domains: FFIW, KoDF, FSh, and DFo.
This result highlights that our modality-specific frequency distillation achieves adaptive knowledge transfer by preserving the most relevant past representations while aligning them with new spatiotemporal cues, thus generalizing robustly to unseen deepfakes.

\subsubsection{Robustness Analysis}
To further assess the robustness of our method, we evaluate the models on FF++ under four common perturbations: H.264 video compression (FF++ C40), resize, grid shuffle, and frame shuffle.
As shown in Table~\ref{tab:perturbation}, while existing continual learning approaches (e.g., iCaRL-FC, DFIL, and SURLID) exhibit notable drops in performance, our model maintains consistently higher performance across all conditions.
In particular, ours performs robustly under both temporal~(frame shuffling) and spatial~(grid shuffling) perturbations, demonstrating its ability to effectively preserve both spatial and temporal information.

\begin{table}[t!]
\centering
\begin{minipage}[t]{0.48\columnwidth}
\centering
\caption{\textbf{Unseen deepfake generalization}.
All models are trained sequentially on datasets up to DFDCp on Protocol 1, and evaluated on unseen forgery datasets (FFIW, KoDF, FSh, DFo).
}
\vspace{-2mm}
\resizebox{\linewidth}{!}{
\begin{tabular}{l|cccc|c}
\toprule
\textbf{Method} & \textbf{FFIW} & \textbf{KoDF} & \textbf{FSh} & \textbf{DFo} & \textbf{Avg. AUC} $\uparrow$ \\
\midrule
Finetune & 66.45 & 85.53 & 69.26 & 82.51 & 75.94 \\
iCaRL-FC~\cite{rebuffi2017icarl} & 69.11 & 84.58 & \underline{79.61}& \textbf{92.66} & \underline{81.49} \\
DFIL~\cite{pan2023dfil} & 68.33 & \textbf{86.89}& 76.03 & 83.65 & 78.73 \\
SURLID~\cite{surlid} & \underline{69.29}& 82.60 & 71.78 & \underline{90.72} & 78.60 \\
\midrule
\textbf{Ours} & \textbf{72.53} & \underline{85.63} & \textbf{86.18}& 90.47 & \textbf{83.70} \\
\bottomrule
\end{tabular}
}
\label{tab:unseen_generalization}
\end{minipage}
\hfill
\begin{minipage}[t]{0.48\columnwidth}
\centering
\caption{\textbf{Perturbation robustness analysis.}
After completing all tasks, we evaluate robustness on FF++ under four perturbations: compression, resize, frame shuffle, and grid shuffle.}
\vspace{-2mm}
\resizebox{\linewidth}{!}{
\begin{tabular}{l|cccc}
\toprule
\textbf{Method} & \textbf{Compression} & \textbf{Resize}  & \textbf{Frame Shuffle} & \textbf{Grid Shuffle} \\
\midrule
Finetune & 67.21 & 73.31  & 70.28 & 68.43 \\
iCaRL-FC~\cite{rebuffi2017icarl} & \underline{72.29} & \underline{79.61}  & \underline{71.22} & 71.23 \\
DFIL~\cite{pan2023dfil} & 70.74 & 77.88  & 64.36 & \underline{72.81} \\
SURLID~\cite{surlid} & 70.54 & 78.07  & 65.87 & 64.94 \\
\midrule
\textbf{Ours} & \textbf{77.61} & \textbf{85.86}  & \textbf{77.71} & \textbf{77.34} \\
\bottomrule
\end{tabular}
}
\label{tab:perturbation}
\end{minipage}
\end{table}

\subsubsection{Scalability to long task sequences}
\begin{table}[t]
\centering
\caption{\textbf{Long-sequence (12-task Fake-Incremental).}
We construct a 12-task long sequence under the Fake-Incremental setting: starting from FF++, we sequentially add 6 DF40~\cite{DF40} forgery types (BlendFace, FaceDancer, MobileSwap, Hyperreenact, MCNet, Sadtalker), followed by 5 deepfake datasets (DFD, CDF, DFDCp, FFIW, KoDF) as fake-only tasks.
}
\label{tab:rebuttal_long}
\begin{tabular}{l|cc}
\toprule
Method & $AUC_\text{all}~\uparrow$ & $F_\text{avg}~\downarrow$\\
\midrule
iCaRL-FC & 86.76 & 3.21 \\
DFIL     & 85.11 & 2.39 \\
SURLID   & 85.16 & 2.79 \\
STSP     & 67.43 & -4.58 \\
\textbf{MSFD (Ours)} & \textbf{87.44} & \textbf{0.85} \\
\bottomrule
\end{tabular}%
\end{table}

We construct a 12-task long sequence under the Fake-Incremental setting to evaluate scalability to long task sequences.
MSFD achieves the highest $AUC_\text{all}$ (87.44) with competitive forgetting (Tab.~\ref{tab:rebuttal_long}), confirming graceful scaling under long-term forgery evolution. STSP attains a lower nominal $F_\text{avg}$ but only by sacrificing plasticity ($AUC_\text{all}=67.43$), failing to adapt to new forgeries; a negative $F_\text{avg}$ here reflects a model that barely learns rather than one that forgets little.

\subsection{Additional Ablation Study}
\subsubsection{Additional Analysis for Effect of the Modality-Specific Adaptor (MSA)}

\begin{figure}[t!]
    \centering
    \begin{subfigure}{\columnwidth}
        \centering
        \includegraphics[width=\columnwidth]{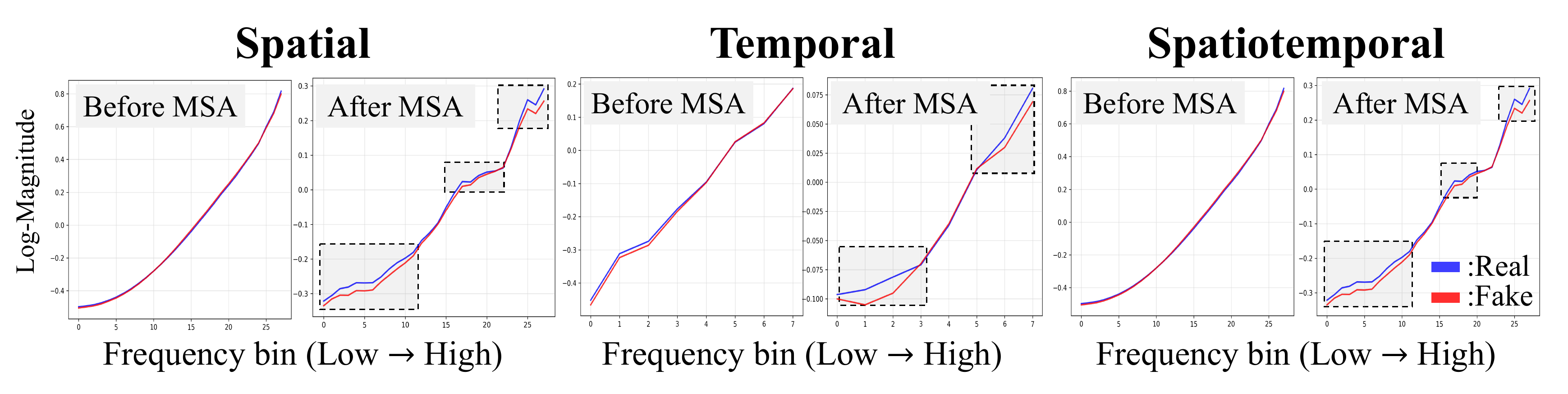}
        \caption{Layer 2}
    \end{subfigure}
    \vspace{2mm}

    \begin{subfigure}{\columnwidth}
        \centering
        \includegraphics[width=\columnwidth]{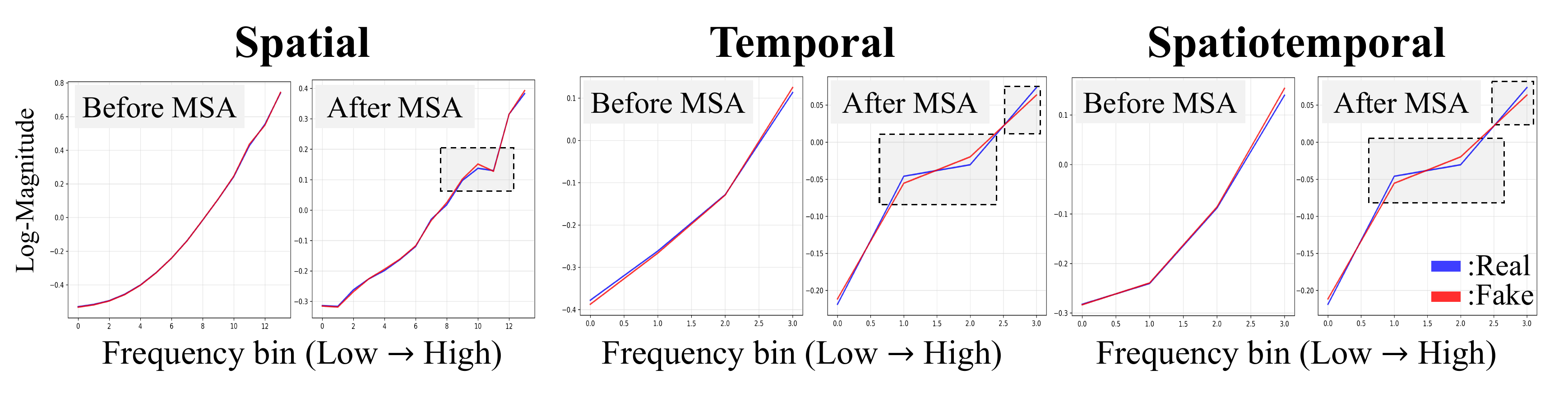}
        \caption{Layer 3}
    \end{subfigure}
    \vspace{2mm}

    \begin{subfigure}{\columnwidth}
        \centering
        \includegraphics[width=\columnwidth]{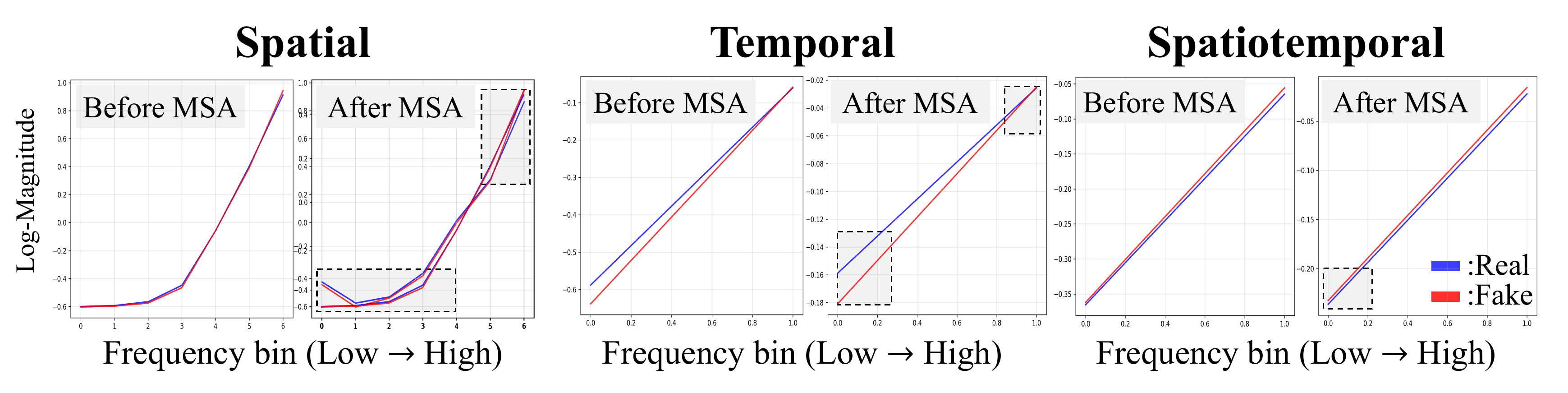}
        \caption{Layer 4}
    \end{subfigure}
    \caption{\textbf{Additional analysis for effect of the Modality-Specific Adaptor (MSA) across Layers 2--4.}
    Frequency spectra of real (blue) and fake (red) samples are visualized across the spatial, temporal, 
    and spatiotemporal modalities for Layers~2 to~4. 
    For each layer, the upper row shows distributions before applying MSA, and the lower row shows distributions after applying MSA. 
    Each distribution is computed using 100 real and 100 fake samples.}
    \label{fig:frequency_all_layers}
\end{figure}

To analyze the impact of MSA, we further visualize the frequency distributions of real and fake samples at layers 2-4, representing features, across spatial, temporal, and spatiotemporal modalities using our model. 
As shown in Fig.~\ref{fig:frequency_all_layers}, each subfigure compares the real–fake frequency spectra before (top) and after (bottom) applying the Modality-Specific Adaptor (MSA) across Layers 2–4.
Before applying MSA, the spectral responses of real and fake videos substantially overlap, indicating limited discriminability.
After applying MSA, however, the separation between real and fake distributions becomes consistently more pronounced in all modalities—spatial, temporal, and spatiotemporal—demonstrating that MSA effectively amplifies forgery-relevant spectral cues at every layer.

\subsubsection{Analysis on Modality Decomposition Efficacy}
To verify whether our framework effectively decomposes video features into meaningful spatial and temporal components, we analyze the performance of individual modality branches: Spatial ($\mathsf{S}$), Temporal ($\mathsf{T}$), and Spatiotemporal ($\mathsf{ST}$).
Each model using only the corresponding single modality is trained without CDL.
Also, we categorize the manipulation types in FF++ into Face Swap (DF, FS), which involves replacing the entire face identity, and Face Reenactment (F2F, NT), which manipulates expressions and movements while maintaining identity.
\begin{table}[t!]
\centering
\caption{
\textbf{Additional analysis for the effect of modality decomposition.}
We analyze the contribution of each frequency modality: spatial ($\mathsf{S}$), temporal ($\mathsf{T}$), and spatiotemporal ($\mathsf{ST}$). 
Results are reported on FF++ with four manipulation types, grouped by their characteristics: face swap deepfakes (DF, FS) and face reenactment deepfakes (F2F, NT). We denote video-level AUC (\%).}
\label{tab:additional_modality}
\resizebox{\columnwidth}{!}{
\begin{tabular}{l|cc|cc|c}
\hline
 & \multicolumn{2}{c|}{\textbf{Face Swap}} & \multicolumn{2}{c|}{\textbf{Face Reenactment}} & \\
\textbf{Model} & DF & FS & F2F & NT & Avg.\\ 
\hline
Spatial Only ($\mathsf{S}$) & 99.88 & 97.95 & 97.79 & 94.93 & 97.64\\
Temporal Only ($\mathsf{T}$) & 99.88 & 97.86 & 98.36 & 95.44 & 97.89\\
Spatiotemporal Only ($\mathsf{ST}$) & 99.95 & 99.42 & 97.94 & 94.94 & 98.06\\
\hline
Ours (Full) & 99.83 & 99.33 & 99.01 & 96.67 & 98.71\\
\hline
\end{tabular}}
\end{table}

As presented in Table~\ref{tab:additional_modality}, the performance trends across different manipulation types provide empirical evidence for the effectiveness of our modality decomposition. 
For Face Swap (DF, FS), which introduces significant spatial and temporal anomalies, every model achieves high accuracy, indicating that artifacts are prominent in both domains.
However, for Face Reenactment (F2F, NT), where temporal consistency (e.g., mouth movement, expression jitter) is a key discriminatory factor, the temporal only model ($\mathsf{T}$) consistently outperforms the spatial only model ($\mathsf{S}$), confirming that our framework effectively decomposes and utilizes spatial and temporal cues.
While single-modality models show limitations in specific categories, our Full Model (Ours) integrates these complementary cues to achieve the highest Average AUC.
This substantial gain validates that the synergy between spatial and temporal representations is essential for robustly detecting sophisticated manipulation types.


\subsubsection{Analysis on Frequency Components}
Frequency representations consist of two components: magnitude and phase. To examine their respective effects, we compare models built on the same overall framework, differing only in the frequency component used for distillation, as shown in Table~\ref{tab:frequencycomponennt}.
\begin{wrapfigure}{r!}{0.48\columnwidth}
  \centering
\vspace{-9mm}
    \includegraphics[width=\linewidth]{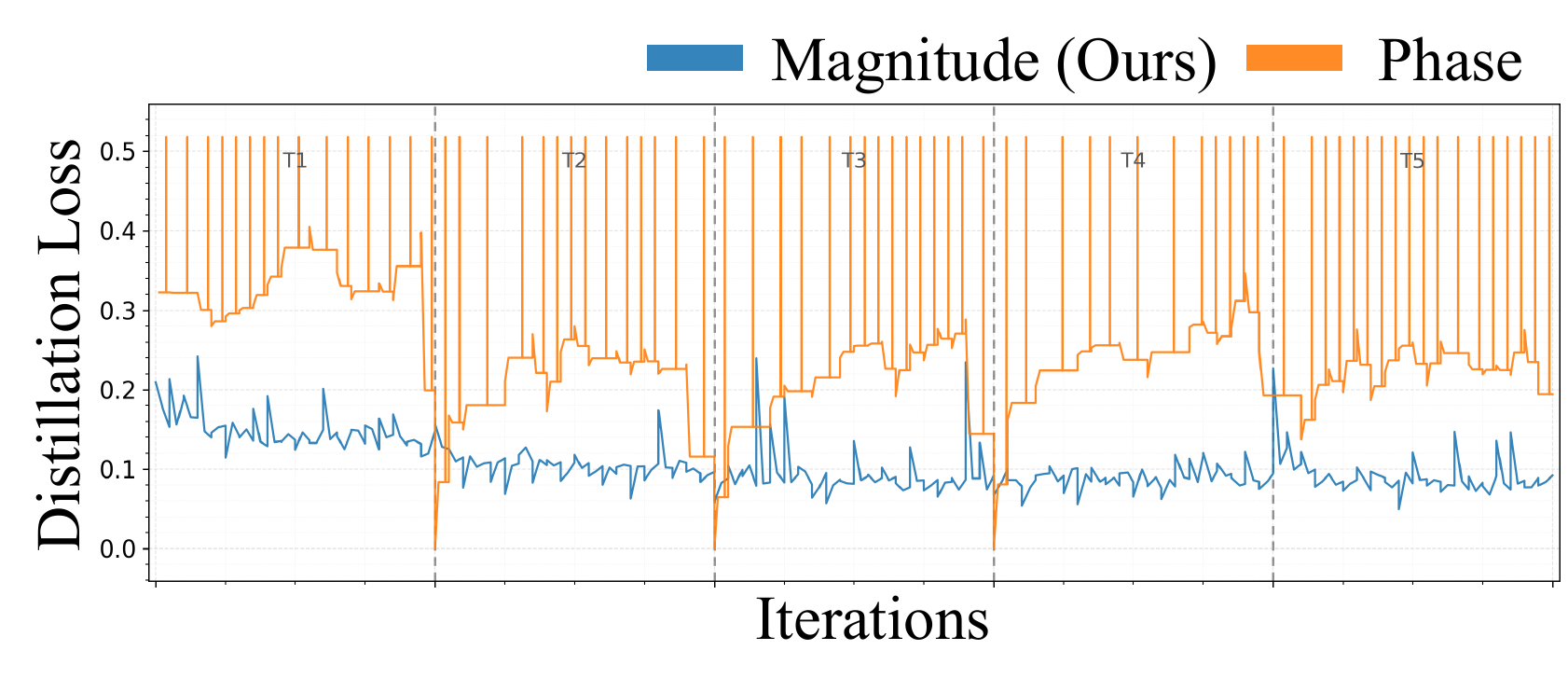}
  \caption{\textbf{Visualization distillation loss}. 
  We compare the feature distillation loss $\mathcal{L}_{FDL}$ when using phase and magnitude components.}
  \label{fig:lossphase}
  \vspace{-9mm}
\end{wrapfigure}
The results show that using magnitude leads to better continual deepfake detection performance than using phase, achieving higher $AUC_{all}$ and lower average forgetting $F_{avg}$. 
In contrast, when phase is used, the training process frequently becomes unstable during continual updates, and the distillation loss often diverges, eventually producing NaN values, as illustrated in Fig.~\ref{fig:lossphase}.
We believe this is because the phase component is highly sensitive to small changes in the input and can vary abruptly across tasks, making it less stable for preserving transferable knowledge in continual learning. 
These observations support our choice of magnitude as the frequency representation in the proposed method.

\begin{table}[t!]
    \centering
    \caption{\textbf{Analysis of frequency component.} We analyze the frequency components, including phase and magnitude. }
    \label{tab:frequencycomponennt}
    \begin{tabular}{c|cc}
        \hline
         \textbf{Component} & \textbf{$AUC_{all}\uparrow$}  & \textbf{$F_{avg}\downarrow$} \\ \hline
          Phase & 90.04 & 5.53\\
         Magnitude (Ours) & 93.02 & 2.29 \\ \hline
    \end{tabular}
\end{table}

\subsubsection{Analysis of Memory Update Strategy}
We analyze the effect of different memory update strategies within our framework, including random replacement and class-center based update, as shown in Table~\ref{tab:replay}. The overall performance remains largely consistent across strategies, with only marginal differences in $AUC_{all}$, suggesting that our framework is stable with respect to the choice of memory update policy.

Nevertheless, the difference in average forgetting $F_{avg}$ is more pronounced. Although random replacement achieves comparable $AUC_{all}$, it leads to higher forgetting than the center-based update. This implies that the memory update strategy plays an important role in preserving previously acquired knowledge over time. In particular, our results suggest that future memory construction schemes may benefit from explicitly considering the relative importance of spatial and temporal cues, rather than treating all samples equally during memory update.

\begin{table}[t!]
    \centering
    \caption{\textbf{Analysis of memory update strategy.} We evaluate different memory update strategies in our framework. }
    \label{tab:replay}
    \begin{tabular}{c|cc}
        \hline
         \textbf{Strategy} & \textbf{$AUC_{all}\uparrow$}  & \textbf{$F_{avg}\downarrow$} \\ \hline
          Random~\cite{er} &  93.01 &	3.89 \\
             Center (default)~\cite{rebuffi2017icarl}&   93.02	& 2.29 \\ \hline
    \end{tabular}
\end{table}

\begin{table}[t!]
\centering
\caption{\textbf{Further analysis of modality decomposition.} We compare
MSFD against a simpler convolutional alternative and an FFT-free variant
under Protocol~1. \emph{Conv.\ Decomposition} replaces the frequency
transform with per-modality 3D convolutions; \emph{MSFD w/o FFT} removes
the FFT while keeping MSA and CDL. We report video-level $AUC_\text{all}$
(\%) and average forgetting $F_\text{avg}$.}
\label{tab:rebuttal_ablation_nofft}
\begin{tabular}{l|cc}
\toprule
Variant & $AUC_\text{all}$~$\uparrow$ & $F_\text{avg}$~$\downarrow$ \\
\midrule
Multi-layer Distill.\ (raw feature) & 90.64 & 6.34 \\
Conv.\ Decomposition (3D conv)      & 91.64 & 4.64 \\
MSFD w/o FFT                        & 92.21 & 3.87 \\
\midrule
\textbf{Full MSFD (Ours)}           & \textbf{93.02} & \textbf{2.29} \\
\bottomrule
\end{tabular}
\end{table}

\subsubsection{Further Analysis on Modality Decomposition}
To isolate the contribution of the frequency transform itself, we add two variants (Tab.~\ref{tab:rebuttal_ablation_nofft}): (i) \emph{MSFD w/o FFT}, a deeper ablation that removes FFT from MSFD while keeping other components (e.g., MSA, CDL); and (ii) \emph{Conv.\ Decomp.}, a simpler alternative that replaces frequency decomposition with per-modality 3D convolutions ($1\!\times\!3\!\times\!3$ / $3\!\times\!1\!\times\!1$ / $3\!\times\!3\!\times\!3$ for $\mathsf{S}$/$\mathsf{T}$/$\mathsf{ST}$).
Both fall short of MSFD, suggesting frequency-domain decomposition contributes beyond axis-separation alone.

\begin{table}[t!]
\centering
\caption{\textbf{Robustness to frequency-suppressing perturbations.}
After completing all tasks under Protocol~1, we report video-level AUC (\%) on FF++ under three perturbations that attenuate high-frequency forensic cues: Gaussian blur, WebP compression, and a 3D-FFT low-pass filter (cutoff $=0.5$). }
\label{tab:freq_robust}
\begin{tabular}{l|ccc}
\toprule
Method & Gaussian Blur & WebP & 3D-FFT Low-pass \\
\midrule
iCaRL-FC          & 47.4 & 77.2 & 71.3 \\
DFIL              & 53.3 & 66.6 & 66.5 \\ \hline
\textbf{MSFD (Ours)} & \textbf{74.9} & \textbf{85.5} & \textbf{74.5} \\
\bottomrule
\end{tabular}
\end{table}

\subsubsection{Robustness to Frequency-Suppressing Perturbations}
We apply three frequency-suppressing perturbations on FF++ (Gaussian blur, WebP, 3D-FFT low-pass cutoff): MSFD retains substantially higher AUC than both iCaRL and DFIL in every case, confirming our frequency preservation provides complementary cues robust to frequency-domain manipulation.

\begin{table*}[t!]
\centering
\caption{\textbf{Comparison with static video-based deepfake detection approaches.} 
Results without a marker are quoted from the original papers, whereas $^*$ indicates reproduced results under a different evaluation setting.}
\label{tab:rebuttal_non-cl}
\begin{tabular}{l|c|c|cc}\hline
\textbf{Method} &Architecture& Params.& CDF&FFIW\\ \hline
 STA~\cite{STa} (CVPR 2025)& CLIP-B16 + Adapter& 106M& 86.6&-\\
 PwTF~\cite{pwtf} (ICCV 2025)& FTCN + Transformer& 164M& 89.7&75.2$^*$\\ \hline
Ours(MSFD)& 3DResNet-18& 33M & 90.2& 94.9\\  \hline
\end{tabular}
\end{table*}

\begin{table}[t!]
\centering
\caption{\textbf{Comparison with deepfake video detectors (sequential FT).} }
\label{tab:rebuttal_comparison}
\resizebox{\columnwidth}{!}{%
\begin{tabular}{l|cccccc|cc}
\toprule
Method & FF++ & DFD & CDF & DFDCp & FFIW & KoDF & $AUC_{all}\uparrow$& $F_\text{avg}\downarrow$\\
\midrule
AltFreezing~\cite{altfreezing} (seq.\ FT)& 62.97 & 78.38 & 58.10 & 62.00 & 57.25 & 98.71 & 69.57 & 27.11 \\ 
PwTF~\cite{pwtf} (seq.\ FT)& 97.28& 84.80& 85.26& 76.01& 52.95&  97.93& 82.37& 9.27\\ 
\midrule
Ours        & 98.42& 91.15 & 90.17 & 84.14 & 94.88 & 99.36 & 93.02 & 2.29 \\
\bottomrule
\end{tabular}%
}
\end{table}

\subsubsection{Comparison with Static Deepfake Video Detection Approaches} 
Our continual learning framework achieves competitive detection performance with $\approx5 \times$ fewer parameters compared to static video-based detectors (Table~\ref{tab:rebuttal_non-cl}). This highlights the practical advantage of continual adaptation: rather than training a single large model on all forgery types at once, MSFD incrementally absorbs new forgery patterns while preserving previously learned cues through modality-wise distillation.
Tab.~\ref{tab:rebuttal_comparison} verifies the necessity: the strongest static spatiotemporal-disentangled detectors AltFreezing~\cite{altfreezing} and PwTF~\cite{pwtf} collapse under sequential fine-tuning ($F_\text{avg}$ of $27.11$ and $9.27$), confirming that decomposition-for-detection does not transfer to continual learning.

\subsubsection{\textbf{Computational Overhead}}
We report deployment and adaptation costs in Tab.~\ref{tab:rebuttal_overhead}.
\begin{wraptable}{r}{0.5\columnwidth}
\centering
\vspace{-9mm}
\caption{\textbf{Computational \& deployment overhead.} We report deploy-time cost (FLOPs, clips/sec), training cost (memory, number of knowledge distillation (KD) layers), and overall performance ($AUC_{all}$) on Protocol 1.}
\label{tab:rebuttal_overhead}
\resizebox{0.5\columnwidth}{!}{%
\begin{tabular}{l|cc|cc|c}
\toprule
\multirow{2}{*}{Method} & \multicolumn{2}{c|}{\textbf{Deploy}} & \multicolumn{2}{c|}{\textbf{Train}} & \multirow{2}{*}{$AUC_{all}\uparrow$}\\
 & FLOPs(G) & Clip/s & Mem(MB) & KD & \\
\midrule
iCaRL-FC & \textbf{325.9} & \textbf{45.5} & 14{,}756 & 0 & 89.84 \\
SURLID   & \textbf{325.9} & 44.3 & \textbf{12{,}869} & 1 & 87.61 \\
DFIL     & \textbf{325.9} & \textbf{45.5} & 14{,}756 & 1 & 87.88 \\
\textbf{Ours} & \textbf{325.9} & \textbf{45.5} & 28{,}287 & \textbf{4} & \textbf{93.02} \\
\bottomrule
\end{tabular}%
}
\vspace{-4mm}
\end{wraptable}
While MSFD consumes more training memory than baselines, this is a necessary consequence of multi-layer modality decomposition (distilling $\{\mathsf{S},\mathsf{T},\mathsf{ST}\}$ at intermediate layers).
Crucially, all auxiliary modules are training-only and discarded after adaptation: the deployed model has identical FLOPs and memory as baselines while achieving the highest $AUC_{all}$. 
Replay storage is also identical to other memory-based methods (iCaRL, DFIL, SURLID).

\section{Additional Implementation Details}

\subsection{Modality-Specific Distillation Mask Generation}
\begin{figure}[t!]
  \centering
\includegraphics[width=\columnwidth]{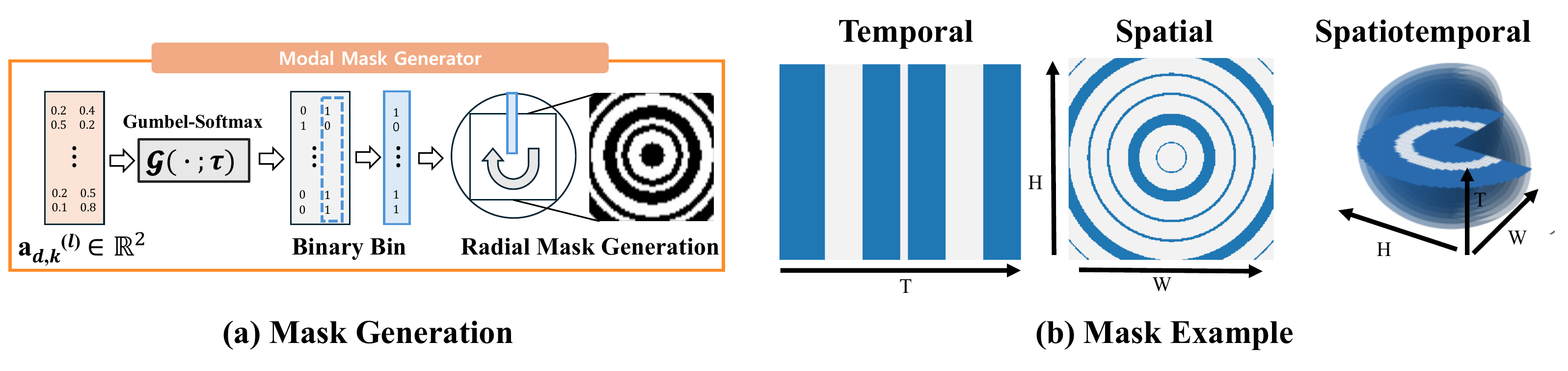}
    \caption{\textbf{Visualization of mask generation.}
(a) Learnable bin-wise logits are transformed into binary pass/block decisions via Gumbel--Softmax and mapped to radial frequency locations to generate the modality-specific distillation mask.
(b) Generated mask examples for the temporal, spatial, and spatiotemporal modalities.}
    \label{fig:maskgeneration}
    \vspace{-5mm}
\end{figure}

As shown in Fig.~\ref{fig:maskgeneration}, we construct a modality-specific radial mask to select informative frequency regions in a data-driven manner. 
For each modality $d \in \{\mathsf{T}, \mathsf{S}, \mathsf{ST}\}$, the radial axis of the corresponding frequency spectrum is discretized into $K_d$ bins, and each bin $k$ is assigned a learnable logit vector $\mathbf{a}_{d,k}^{(l)} \in \mathbb{R}^2$. 
The Gumbel--Softmax operator produces a binary pass/block decision $m_{d,k}^{(l)} \in \{0,1\}$, which is then mapped to radial frequency locations via $\kappa_d^{(l)}(\mathbf{u})$ to form the final mask $M_d^{(l)}(\mathbf{u})$. 
The resulting mask is broadcast to the corresponding modality spectrum and applied element-wise, enabling selective preservation of discriminative frequency regions. 
As illustrated in Fig.~\ref{fig:maskgeneration}-(b), this yields modality-specific mask structures: 1D bands for $\mathsf{T}$, concentric radial bands for $\mathsf{S}$, and spherical frequency shells for $\mathsf{ST}$. 

\subsection{Dataset Configurations}

For continual learning protocols, Task 1 (FF++) uses the original training and test splits. For subsequent tasks (Task 2-6), we sample 200 real and 200 fake videos for training from each dataset. 
Test set sizes are as follows: FaceForensics++ (FF++) with 1400 videos, DFD contains 3,000 videos (2,531 fake), DFDCp contains 758 videos (488 fake), CDF contains 918 videos (510 fake), FFIW~\cite{FFIW} contains 500 videos (250 fake), and KoDF~\cite{Kwon_2021_ICCV} contains 3,000 videos (2,500 fake). CDF denotes the Celeb-DF-V2 dataset. 
For DF40~\cite{DF40}, we use deepfakes sourced from FF++~\cite{faceforensics}.
For KoDF, due to the large size of the original training set, we construct non-overlapping train and validation sets using only the validation portion of the dataset. To ensure balanced evaluation, we randomly sample 500 videos per forgery method from the remaining test videos, excluding those used for training.
For AIGVDBench~\cite{AIGVDBench}, due to the large size of the original training set, we construct non-overlapping train and validation sets through random sampling. Specifically, we first sample 2,000 real videos and 2,000 fake videos per forgery category. From these, 200 samples are allocated for training, while the remaining 1,800 videos are used for validation to ensure a robust evaluation.

\subsection{Perturbation Settings}
\begin{figure*}[t!]
  \centering
\includegraphics[width=\textwidth]{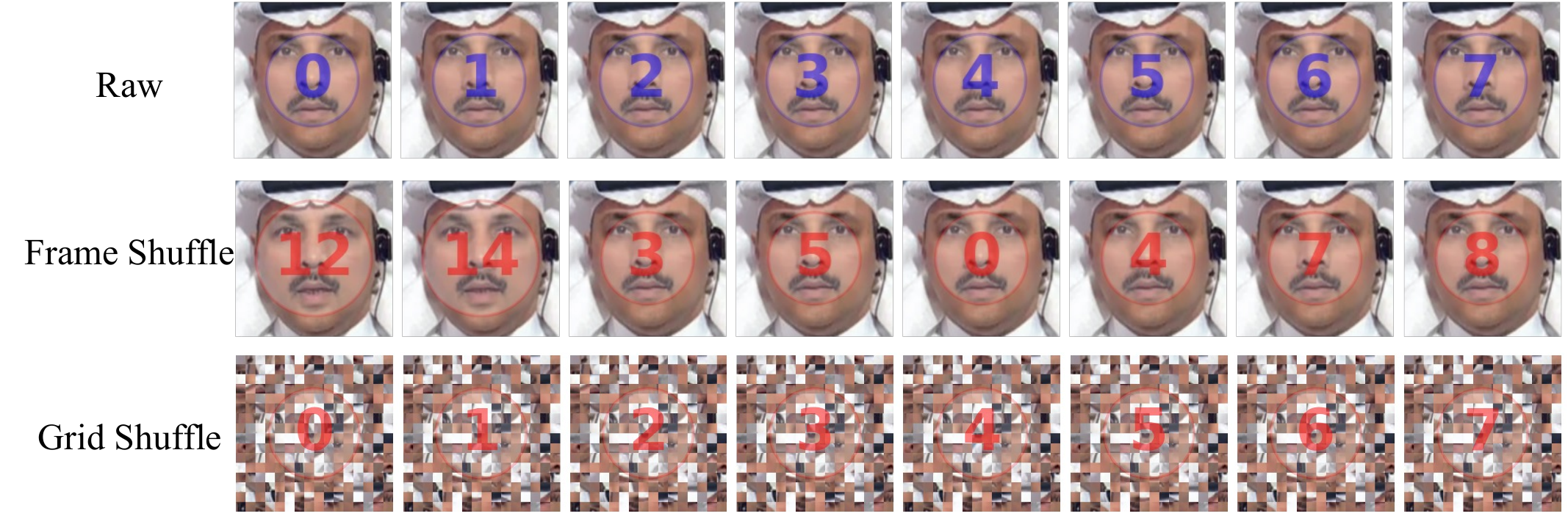}
    \caption{\textbf{Visualization of frame and grid shuffle.} To assess the stability of our learned representations and conduct the preliminary analysis (as discussed in Fig.~2 in the main paper), we apply frame shuffling and grid shuffling to perturb temporal continuity and spatial structure, respectively. The first row shows the raw video clip. The second row shows frame shuffling, where the temporal order is randomized. The third row depicts grid shuffling, where spatial patches are rearranged within each frame using a consistent permutation across the clip while preserving the temporal order. The numbers overlaid on each frame indicate the original frame index.}
    \label{fig:pertur_example}
\end{figure*}

We evaluate the robustness of our model using four perturbation types. First, to simulate real-world quality degradation, we employ \textbf{Video Compression} (using the FF++ C40 dataset) and \textbf{Resizing}, where frames are downsampled to a scale of 0.5 and subsequently upsampled to the original resolution. These perturbations allow us to assess model stability under common transmission artifacts and reduced resolutions. Secondly, to investigate the model's reliance on specific modalities, we apply structural perturbations, including grid shuffle and frame shuffle, as visualized in Fig.~\ref{fig:pertur_example}. \textbf{Grid Shuffle} rearranges local patches within frames using a $G \times G$ grid (we set $G=14$, yielding $16 \times 16$ pixel patches at $224 \times 224$ resolution) to break global spatial structure, whereas \textbf{Frame Shuffle} groups every $K$ consecutive frames into segments and randomizes the segment order (we set $K=1$, \ie, individual frames are fully shuffled) to disrupt motion continuity. For the visualization in Fig. 2, we use $G=7$ and $K=2$ for clearer visual illustration. These settings allow us to effectively decouple the contribution of spatial and temporal features under adversarial conditions.


To investigate how continual learning affects spatial and temporal knowledge independently, we measure modality-specific retention after all tasks have been learned sequentially (Fig.~2, Fig.~6).
Specifically, we define spatial retention as the ratio of the final model's AUC under grid-shuffle perturbation to its AUC when the dataset was first encountered, and temporal retention analogously under frame-shuffle perturbation.
A value closer to 1.0 indicates that the model preserves its original discriminative capability in that modality despite subsequent task updates.

\subsection{Training Configuration} 
During training, all frames in a clip share the same augmentation parameters to preserve temporal coherence. Each frame is first resized to an intermediate scale $s \sim [256, 320]$, followed by a random crop with size $c \sim [\text{frame\_size}, s]$. A horizontal flip with probability 0.5 is then applied at the clip level. We additionally apply stochastic masking with up to three holes, where the generated binary mask is broadcast across channels and multiplied with each frame. After augmentation, frames are resized to $224 \times 224$ and normalized using ImageNet~\cite{ImageNet} statistics (mean: $(0.485, 0.456, 0.406)$; std: $(0.229, 0.224, 0.225)$). The processed frames are then stacked along the temporal dimension to form a clip tensor of size $[C, T, H, W]$, where $C=3$, $T=32$, and $H=W=224$. At evaluation time, only resizing and normalization are applied.
In the R3D-18~\cite{3dresnet} backbone, which we generally adopt, the number of layers used for feature extraction is $L=4$. 
The loss weighting coefficients are set as $\lambda_{\text{PD}}$ is 1.0, $\lambda_{\text{FDL}}$ is 1.0, and $\lambda_{\text{CDL}}$ is 0.5.
For modality-specific frequency distillation, the weights are $\lambda_{S}$ is 0.5, $\lambda_{T}$ is 1.0, and $\lambda_{ST}$ is 1.0. 

Most experiments were carried out on either four NVIDIA A6000 GPUs (48GB each) paired with an AMD Ryzen Threadripper Pro 3955WX 16-core CPU, or on two NVIDIA A100 GPUs (40GB each) coupled with an Intel Xeon Gold 6240R CPU. Training on Protocol 1 takes approximately 30 minutes per epoch using A100 GPUs.

\subsection{Configuration of Comparison Methods}
We provide detailed implementation information for each continual learning method.

For the Nearest Mean of Exemplars classifier (NME), the probability of a sample $x$ being fake is computed as:
\begin{equation}
p_{\text{fake}}(x)
= \sigma\!\left(d_{\text{real}} - d_{\text{fake}}\right),
\label{eq:nme_prob}
\end{equation}
where the distances to the class prototypes are defined as $d_{\text{cls}} = \| f - \mu_{\text{cls}} \|_2, \text{cls}\in \{real,fake\}$, and
$f = \phi(x), \quad \sigma(z) = \frac{1}{1 + e^{-z}}$.
Here, $f$ denotes the L2-normalized feature vector extracted from the backbone $\theta$,
$\mu_{\text{real}}$ and $\mu_{\text{fake}}$ represent the class mean prototypes computed from exemplars,
$d_{\text{real}}$ and $d_{\text{fake}}$ are their Euclidean distances,
and $\sigma(\cdot)$ is the sigmoid function.

\noindent\textbf{ER~\cite{er}.}
We reproduce ER~\cite{er} following the original paper.
The method maintains a memory buffer populated with randomly selected exemplars, where class balance is preserved via simple class-wise pruning.
During continual training, current-task samples are mixed with replayed exemplars to form each minibatch, and the model is optimized solely with standard classification loss on this combined batch.
No additional regularization, distillation, or task-specific constraints are introduced, ensuring a faithful reproduction of the baseline ER protocol.

\noindent\textbf{EWC~\cite{ewc}.}
We reproduce EWC~\cite{ewc} following the original paper.
After completing each task, we estimate the Fisher Information Matrix using samples from the current task and accumulate it across tasks to model parameter importance.
The quadratic regularization term is then applied during training to penalize deviations from important parameters of previous tasks.
Following common practice, we set the regularization coefficient to $\lambda = 1000$ for all experiments.

\noindent\textbf{iCaRL~\cite{rebuffi2017icarl}.}
We reproduce iCaRL~\cite{rebuffi2017icarl} using herding-based exemplar selection as described in the original work.
During training, the model is optimized using a combination of classification loss on mixed current–memory batches and a distillation loss computed on memory samples, both weighted equally at 1.0.
For evaluation, we report results using both the Nearest Mean of Exemplars (NME) classifier, computed with exemplar-based class means described in Equation~\ref {eq:nme_prob}, and the standard Fully-Connected (FC) linear classifier

\noindent\textbf{IDER~\cite{liu2026ider}.}
We reproduce IDER~\cite{liu2026ider} following their official implementation, adapting its idempotent experience replay to the binary deepfake setting.
The R3D-18 backbone is split into a feature stage $f_1$ and a label-prior-conditioned prediction stage $f_2$, and IDER enforces idempotency by applying $f_2$ twice with cross-entropy on both passes so the prediction converges to a stable fixed point.

\noindent\textbf{DCE~\cite{li2025addressing}.}
We adapt the domain-incremental learning method DCE~\cite{li2025addressing} to continual deepfake video detection following their official implementation.
DCE expands a pool of three collaborative experts per task: a naive expert trained with plain cross-entropy, together with balanced and reverse experts whose objectives add the log class prior once and twice, so the three heads specialize to the natural, balanced, and minority-emphasizing views of the same features.
Following the original setup, the backbone is shared and frozen after the first task, and at each subsequent task, only the current task's three experts are updated while all earlier experts remain fixed.
After expert training, we estimate per-class Gaussian prototypes and train a dynamic selector on synthetic features sampled from these prototypes with task-distance-dependent scaling, so no raw exemplars are retained; at inference the selector produces per-expert weights that fuse all experts' logits into the final real--fake decision.

\noindent\textbf{vCLIMB~\cite{vclimb}.}
We reproduce vCLIMB~\cite{vclimb} following their official implementation code.
The method extends iCaRL replay with a temporal consistency regularization that enforces prediction invariance between original and temporally downsampled clips.
During training, each clip is paired with a temporally downsampled view reconstructed via trilinear interpolation, and the final loss blends their classification objectives.
We adopt herding-based exemplar selection and report results using NME classifiers.

\noindent\textbf{Drift~\cite{drift}.} 
We adapt the video domain-incremental learning method Drift~\cite{drift} to the deepfake video detection setting.
Drift employs a dual-loss formulation that combines standard classification loss on both current and replayed samples with a knowledge distillation term to preserve past representations.
Following the original setup, we set the distillation weight to 1.0 and the temperature to 2.0.

\noindent\textbf{HCE~\cite{HCE}.} 
We compare the conventional video class-incremental learning method HCE~\cite{HCE} with our approach for continual deepfake video detection.
HCE computes spatial hypercorrelation matrices between intermediate features of the current and previous models across backbone layers, aggregates them through 4D convolutions, and uses the resulting hypercorrelation weights to modulate the feature-level distillation objective, following the original setting where both distillation and hypercorrelation supervision terms are given equal importance.
In addition, the Correlation Refinement Mechanism (CRM) adjusts the cross-entropy loss based on class-wise hypercorrelation variance ratios, with CRM contributing half the weight of the main classification term to address distribution shifts between old and new data.
For memory construction, we adopt herding-based exemplar selection, and inference follows the original design using an NME classifier for real–fake prediction.

\noindent\textbf{STSP~\cite{stsp}.}
We reproduce STSP~\cite{stsp} following the original paper.
The method learns compact subspace bases per class via a Task-Specific Classifier (TSC), where each class is represented by an orthogonal basis of rank $\tau = 8$ in the feature space.
Classification is performed by comparing subspace distances: the logit is computed as $d_{\text{real}} - d_{\text{fake}}$, where $d_{cls}$ measures the Frobenius distance between the feature's self-correlation and the class subspace projection.
The training objective combines classification loss ($\alpha = 1.0$), a feature orthogonality loss ($\beta = 0.05$), and a weight orthogonality loss ($\gamma = 0.05$) that enforces both intra-class orthonormality and inter-class decorrelation of the subspace bases.
To mitigate forgetting, Subspace Gradient Projection (SGP) constrains gradient updates to the orthogonal complement of previously learned input subspaces, with the energy retention threshold set to 0.99.
SGP is applied to all convolutional and linear layers in the backbone as well as the TSC bases, and no replay buffer is used.

\noindent\textbf{ESSENTIAL~\cite{lee2025essential}.}
We reproduce ESSENTIAL~\cite{lee2025essential} following their official implementation code.
The method builds on a CLIP-based backbone and introduces learnable semantic prompts that capture task-specific knowledge through cross-attention with input features.
Training combines a static matching loss and a temporal matching loss, both weighted equally at 1.0, to align prompt representations with frame-level and sequence-level features.
We set the semantic prompt length to 32 (matching the number of input frames), use cross-attention as the prompt interaction mode, and adopt task-wise memory construction with a buffer size of 200 per task.
Each task is trained for 30 epochs with a learning rate of $1 \times 10^{-4}$.

\noindent\textbf{SARB-DF~\cite{prathibha2024sarb}}. 
We reproduce SARB-DF~\cite{prathibha2024sarb} following the original paper.
For memory update, we use a self-adaptive replay buffer mechanism that dynamically adjusts the composition of stored exemplars based on sample difficulty and representativeness.
The training objective combines standard classification loss with a replay-based regularization term (EWC~\cite{ewc}) to mitigate forgetting across sequential deepfake detection tasks.
We set the total memory buffer size to 500 and follow the original hyperparameter settings for all experiments.

\noindent\textbf{DFIL~\cite{pan2023dfil}}. 
We reproduce DFIL following their official implementation code.
The training objective consists of a classification loss, a supervised contrastive loss ($\tau = 0.1$), feature-level distillation, and a logit-level knowledge distillation term ($T = 20$), each applied with equal weight (1.0).
For memory selection, we adopt DFIL’s hybrid replay strategy: half of the exemplars are chosen based on high entropy (hard samples), while the remaining half are selected according to their proximity to the class centroid (central samples).

\noindent\textbf{HDP~\cite{hdp}}. 
We reproduce HDP~\cite{hdp} following the official implementation. 
Instead of storing raw exemplars, the method utilizes Universal Adversarial Perturbations (UAPs) generated from each task's real videos as compact surrogates for historical distributions. 
Each UAP ($\delta$) is optimized over 10 iterations with a step size of $\eta = 0.01$, targeted to maximize the model's prediction toward the fake class. 
During subsequent tasks, pseudo-fake samples are synthesized on the fly by adding a randomly selected historical UAP to the current real videos. 
The input size is set to $224 \times 224$ with 32 frames per clip. 
Crucially, for our experiments, the UAPs are stored with a temporal dimension of 32 ($1 \times 3 \times 32 \times 224 \times 224$), matching the clip length to ensure precise frame-wise additive synthesis throughout the video sequence.

\noindent\textbf{SURLID~\cite{surlid}.} 
We reproduce SURLID~\cite{surlid} following their official implementation code.
The method introduces \emph{aligned feature isolation} to reduce catastrophic forgetting in incremental deepfake detection and builds on three core components:
(1) Sparse Uniform Replay (SUR), which selects stable and uniformly distributed exemplars based on shuffle consistency and spatial distribution; 
(2) Distribution Refilling (DR), which augments replay features by mixing samples with their domain centroids; and
(3) Incremental Decision Alignment (IDA), which encourages angular consistency between new and previous task-specific classifiers ($\gamma = 0.001$).
Training follows the original SURLID loss, combining isolation loss (supervised contrastive), feature distillation with weight 0.5, and detection loss also weighted by 0.5, which consistently yielded better performance in our continual deepfake video detection experiments.
For memory construction, we adopt SUR sampling, selecting exemplars whose representations remain stable across perturbations (measured using grid shuffling).

\bibliographystyle{splncs04}
\bibliography{main}
\end{document}